\documentclass[11pt,oneside,fleqn,reqno,titlepage]{article}
\usepackage{amsmath}
\usepackage{amssymb}
\usepackage{amsthm}
\usepackage{natbib} \citeindextrue
\usepackage{comment}
\usepackage{url}
\usepackage{bbm}
\usepackage{bm}
\usepackage{multirow}
\usepackage{relsize}
\usepackage[pdftex]{graphicx}
\usepackage{color}

\newcommand{\doublespace}{\addtolength{\baselineskip}{.25\baselineskip}}

\newcommand{\singlespace}{\addtolength{\baselineskip}{-.5\baselineskip}}
\newcommand{\dense}{\renewcommand{\arraystretch}{1}}

\newcounter{stepno}
\newcommand{\horizontalline}{\noindent \mbox{}\hrulefill\mbox{}}

\newcommand{\E}{\mathbb{E}}
\newcommand{\bn}{\begin{eqnarray}}
\newcommand{\en}{\end{eqnarray}}
\newcommand{\bns}{\begin{eqnarray*}}
\newcommand{\ens}{\end{eqnarray*}}

\newcommand{\defvarbegin}{\begin{quotation}\vspace{-15pt}\begin{tabbing}}
\newcommand{\defvarend}  {\end{tabbing}\vspace{-10pt}\end{quotation}}

\newcommand{\bnarray}{\begin{equation}\begin{array}{rcll}}
\newcommand{\enarray}{\end{array}\end{equation}}

\newcommand{\textwrap}{\parbox[t]{5.0in}}

\newcommand{\textwrapsmall}{\parbox[t]{4.5in}}

\newcommand{\barr}{\begin{array}}
\newcommand{\earr}{\end{array}}

\newcounter{cnum}

\newcommand{\beginalg}{\setcounter{stepno}{1}
                \begin{list}{\bf Step~\arabic{stepno}}
                         {\usecounter{stepno}\settowidth{\labelwidth}{\bf Step~9m}
                \addtolength{\leftmargin}{2\parindent}}
                }
\newcommand{\eg}{\end{list}}

\newcommand{\remark}{{\noindent \bf Remark:}~}

\def \define{\begin{quote}\begin{itemize}}
\def \enddefine{\end{itemize}\end{quote}}

\newlength{\boxedparwidth} 
  {\begin{center} \begin{tabular}{|@{\hspace{.15in}}c@{\hspace{.15in}}|}
                 \hline \\ \begin{minipage}[t]{\boxedparwidth}
                 \setlength{\parindent}{0.0in}}%
   {\end{minipage} \\ \\ \hline \end{tabular} \end{center}}
\newcounter{example}
\def \rhooperator{\mathlarger{\mathlarger{\rho}}}

\newcommand{\argmax}{{\rm arg}\max}

\def \xtilde{{\tilde x}}

\def \Etilde{{\tilde E}}

\def \Stilde{{\tilde S}}

\def \Vtilde{\stackrel{\sim\;}{V}}

\def \Wtilde{{\tilde W}}
\def \Xtilde{{\tilde X}}

\def \phat{\hat p}

\def \Bhat{\hat B}

\def \Fhat{\hat F}

\def \What{\widehat W}

\def \thetabar{\bar \theta}
\def \deltabar{\bar \delta}
\def \deltahat{\hat \delta}

\def \hbar{\bar h}

\def \pbar{\bar p}

\def \Cbar{\bar C}

\def \Fbar{\bar F}
\def \Qbar{\bar Q}

\def \Vbar{\bar V}

\def \mubar{\bar \mu}
\def \sigmabar{\bar \sigma}

\def \Acal{{\cal A}}

\def \Fcal{{\cal F}}

\def \Ical{{\cal I}}

\def \Ocal{{\cal O}}
\def \Pcal{{\cal P}}
\def \Qcal{{\cal Q}}

\def \Scal{{\cal S}}

\def \Wcal{{\cal W}}
\def \Xcal{{\cal X}}
\def \Ycal{{\cal Y}}
\def \Zcal{{\cal Z}}

\begin{document}

\title{On State Variables, Bandit Problems and POMDPs}
\author{Warren B. Powell \\ Department of Operations Research and Financial Engineering \\ Princeton University \\}
\date{\today}

\maketitle

\clearpage

\begin{abstract}
State variables are easily the most subtle dimension of sequential decision problems.  This is especially true in the context of active learning problems (``bandit problems'') where decisions affect what we observe and learn.  We describe our canonical framework that models {\it any} sequential decision problem, and present our definition of state variables that allows us to claim: Any properly modeled sequential decision problem is Markovian.  We then present a novel two-agent perspective of partially observable Markov decision problems (POMDPs) that allows us to then claim: Any model of a real decision problem is (possibly) non-Markovian.  We illustrate these perspectives using the context of observing and treating flu in a population, and provide examples of all four classes of policies in this setting.  We close with an indication of how to extend this thinking to multiagent problems.
\end{abstract}

\clearpage
\pagestyle{empty}
{\singlespace
\tableofcontents
}
\clearpage

\setcounter{page}{1}
\pagestyle{plain}

\doublespace

\section{Introduction}
Sequential decision problems span a genuinely vast range of applications including engineering, business, economics, finance, health, transportation, and energy.  It encompasses active learning problems that arise in the experimental sciences, medical decision making, e-commerce, and sports.  It also includes iterative algorithms for stochastic search, as well as two-agent games and multiagent systems. In fact, we might claim that virtually any human enterprise will include instances of sequential decision problems.

Sequential decision problems consist of sequences: decision, information, decision, information, $\ldots$, where decisions are determined according to a rule or function that we call a {\it policy}, that is a mapping from a state to a decision. This is arguably the richest problem class in modern data analytics.  Yet, unlike fields such as machine learning or deterministic optimization, they lack a canonical modeling framework that is broadly used.

The core challenge in modeling sequential decision problems is modeling the state variable.  A consistent theme across what I have been calling the ``jungle of stochastic optimization'' (see \url{jungle.princeton.edu}) is the lack of a standard definition (with the notable exception of the field of optimal control).

I am using this document to a) offer my own definition of a state variable (taken from \cite{PowellRLSO2020}), b) suggest a new perspective on the modeling of learning problems (spanning bandit problems to partially observable Markov decision problems) and c) extending these ideas to multi-agent systems.  This discussion draws from several sources I have written recently: \cite{PowellRLSO2020}, \cite{PowellRLHandbook2019}, and \cite{PowellEJOR2019}.  These are all available at jungle.princeton.edu.

In this article, I am going to argue:
\begin{itemize}
  \item[1)] All properly modeled problems are Markovian.
  \item[2)] All models of real applications are (possibly) non-Markovian.
\end{itemize}
These seem to be contradictory claims, but I am going to show that they reflect the different perspectives that lead to each claim.

We are going to begin by presenting our universal framework for modeling sequential decision problems in section \ref{sec:modelingsequentialdecisions}. The framework presents sequential decision problems in terms of optimizing over policies.  Section \ref{sec:designingpolicies} provides a streamlined presentation of how to design policies for any sequential decision problem.

Section \ref{sec:statevariables} then provides an in-depth discussion of state variables, including a brief history of state variables, our attempt at a proper definition, followed by illustrations in a variety of settings.   Then, section \ref{sec:POMDP} discusses  partially observable Markov decision problems, and presents a two-agent model that offers a fresh perspective of all learning problems.  We illustrate these ideas in section \ref{sec:learningflu} using a problem setting of learning how to mitigate the spread of flu in a population.  We then extend this thinking in section \ref{sec:multiagent} to the field of multiagent systems.

This article is being published on arXiv only, so it is not subject to peer review.  Instead,
readers are invited to comment on this discussion at \url{http://tinyurl.com/statevariablediscussion}.

\section{Modeling sequential decision problems}
\label{sec:modelingsequentialdecisions}
Any sequential decision problem can be written as a sequence of state, decision, information, state, decision, information.  Written over time this would be given by
\bns
(S_0, x_0, W_1, S_1, x_1, W_2, \ldots, S_t, x_t, W_{t+1}, \ldots, S_T)
\ens
where $S_t$ is the ``state'' (to be defined below) at time $t$, $x_t$ is the decision made at time $t$ (using the information in $S_t$), and then $W_{t+1}$ is the information that arrives between $t$ and $t+1$ (which is not known when we make the decision $x_t$).  Note that we start at time $t=0$, and assume a finite horizon $t=T$ (standard in many communities, but not Markov decision processes).

There are many problems where it is more natural to use a counter $n$ (as in $nth$ event or $nth$ iteration).  We write this as
\bns
(S^0, x^0, W^1, S^1, x^1, W^2, \ldots, S^n, x^n, W^{n+1}, \ldots, S^N).
\ens
Finally, there are times where we are iterating over simulations (e.g. the $nth$ pass over a week-long simulation of ad-clicks), which we would write using
\bns
(S^n_t,x^n_t,W^n_{t+1})_{t=0}^T \mbox{~~for $n=0, \ldots, N$}.
\ens

The classical modeling framework used for Markov decision processes is to specify the tuple $(\Scal, \Acal, P, r)$ where $\Scal$ is the state space, $\Acal$ is the action space (the MDP community uses $a$ for action), $P$ is the one-step transition matrix with element $p(s'|s,a)$ which is the probability we transition from state $s$ to $s'$ when we take action $a$, and $r = r(s,a)$ is the reward if we are in state $s$ and take action $a$ (see \cite{Puterman05}[Chapter 3]).  This modeling framework has been adopted by the reinforcement learning community, but in \cite{PowellRLHandbook2019} we argue that it does not provide a useful model, and ignores important elements of a real model.

Below we describe the five elements of any sequential decision problem.  We first presented this modeling style in \cite{PowellADP2011}, but as noted in \cite{PowellRLHandbook2019}, this framework is very close to the style used in the optimal control community (see, for example, \cite{lewis2012}, \cite{Ki98} and \cite{Sontag1998}).  After this, we illustrate the framework with a classical inventory problem (motivated by energy storage) and as a pure learning problem.  The framework involves optimizing over policies, so we close with a discussion of designing policies.

\subsection{Elements of a sequential decision problem}
\label{sec:elementssequentialdecision problem}
There are five dimensions of any sequential decision problem: state variables, decision variables, exogenous information processes, the transition function and the objective function.
\begin{description}
  \item[State variables] - The state $S_t$ of the system at time $t$ contains all the information that is necessary and sufficient to compute costs/rewards, constraints, and the transition function (we return to state variables in section \ref{sec:statevariables}).
  \item[Decision variables] - Standard notation for decisions might be $a_t$ for action, $u_t$ for control, or $x_t$, which is the notation we use since it is standard in math programming.  $x_t$ may be binary, one of a finite discrete set, or a continuous or discrete vector.  We let $\Xcal_t$ be the feasible region for $x_t$, where $\Xcal_t$ may depend on $S_t$.  We address the problem of designing the policy to later.

      Decisions are made with a decision function or {\it policy}, which we denote by $X^\pi(S_t)$  where ``$\pi$'' carries the information about the type of function $f\in\Fcal$, and any tunable parameters $\theta \in \Theta^f$.  We require that the policy satisfy $X^\pi(S_t) \in \Xcal_t$.
  \item[Exogenous information] - We let $W_{t+1}$ be any new information that first becomes known at time $t+1$ (we can think of this as information arriving between $t$ and $t+1$).  $W_{t+1}$ may depend on the state $S_t$ and/or the decision $x_t$, so it is useful to think of it as the function $W_{t+1}(S_t,x_t)$, but we write $W_{t+1}$ for compactness.  This indexing style means any variable indexed by $t$ is known at time $t$.
  \item[Transition function] - We denote the transition function by
      \bn
      S_{t+1} = S^M(S_t,x_t,W_{t+1}), \label{eq:transition}
      \en
      where $S^M(\cdot)$ is also known by names such as system model, state equation, plant model, plant equation and transfer function.  $S^M(\cdot)$ contains the equations for updating each element of $S_t$.
  \item[Objective functions] - There are a number of ways to write objective functions.  We begin by making the distinction between state-independent problems, and state-dependent problems.  We let $F(x,W)$ denote a state-independent problem, where we assume that neither the objective function $F(x,W)$, nor any constraints, depends on dynamic information captured in the state variable.  We let $C(S,x)$ capture state-dependent problems, where the objective function (and/or constraints) may depend on dynamic information.

      We next make the distinction between optimizing the cumulative reward versus the final reward.  Optimizing cumulative rewards typically arises when we are solving problems in the field where we are actually experiencing the effect of a decision, whereas final reward problems typically arise in laboratory environments.

      Below we list the objectives that are most relevant to our discussion:
      \begin{description}
      \item[State-independent, final reward] This is the classical stochastic search problem.  Here we go through a learning/training process to find a final design/decision $x^{\pi,N}$, where $\pi$ is our search policy (or algorithm), and $N$ is the budget.  We then have to test the performance of the policy by simulating $\What$ using
          \bn
          \max_\pi \E_{S^0} \E_{W^1, \ldots, W^N|S^0} \E_{\What|S^0} F(x^{\pi,N},\What), \label{eq:stateindependentfinalreward}
          \en
          where $x^{\pi,N}$ depends on $S^0$ and the experiments $W^1, \ldots, W^N$, and where $\What$ represents the process of testing the design $x^{\pi,N}$.
      \item[State-independent, cumulative reward] This is the standard representation of multi-armed bandit problems, where $S^n$ is what we believe about $\E F(x,W)$ after $n$ experiments.  The objective function is written
          \bn
          \max_\pi \E_{S^0} \E_{W^1, \ldots, W^N|S^0} \sum_{n=0}^{N-1}F(X^{\pi}(S^n),W^{n+1}), \label{eq:stateindependentcumulativereward}
          \en
          where $F(X^{\pi}(S^n),W^{n+1})$ is our performance for the $n+1st$ experiment using experimental settings $x^n = X^{\pi}(S^n)$, chosen using our belief $S^n=B^n$ based on what we know from experiments $1, \ldots, n$.
      \item[State-dependent, cumulative reward] This is the version of the objective function that is most widely used in stochastic optimal control (as well as Markov decision processes).  We switch back to time-indexing here since these problems are often evolving over time (but not always).  We write the contribution in the form $C(S_t,x_t,W_{t+1})$ to help with the comparison to $F(x,W)$, which gives us
          \bn
          \hspace{-.2in}\max_\pi \E_{S_0} \E_{W_1, \ldots, W_T|S_0}  \left\{\sum_{t=0}^T C(S_t,X^\pi(S_t),W_{t+1})|S_0\right\}. \label{eq:statedependentcumulativereward}
          \en
      \end{description}
This is not an exhaustive list of objectives (for example, we did not list state-dependent, final reward).  Other popular choices model regret or posterior-optimal solutions to compare against a benchmark.  Risk is also an important issue.

%
\end{description}

Note that the objectives in \eqref{eq:stateindependentfinalreward} -  \eqref{eq:statedependentcumulativereward} all involve searching over policies.  Writing the model, and then designing an algorithm to solve the model, is absolutely standard in deterministic optimization.  Oddly, the communities that address sequential decision problems tend to first choose a solution approach (what we call the policy), and then model the problem around the class of policy.

Our framework applies to {\it any} sequential decision problem.  However, it is critical to create the model before we choose a policy.  We refer to this style as ``{\it model first, then solve}.''  We address the problem of designing policies in section \ref{sec:designingpolicies}.

\subsection{Energy storage illustration}
\label{sec:energystorage}
We are going to use a simple energy storage problem to provide a basic illustration of the core framework.  Our problem involves a storage device (such as a large battery) that can be used to buy/sell energy from/to the grid at a price that varies over time.

\begin{description}
  \item[State variables] State $S_t = (R_t, p_t)$ where
  \bns
  R_t &=& \textwrap{energy in the battery at time $t$,}\\
  p_t &=& \textwrap{price of energy on the grid at time $t$.}
  \ens
  \item[Decision variables] $x_t$ is the amount of energy to purchase from the grid ($x_t > 0$) or sell to the grid ($x_t < 0$).
  We introduce the policy (function) $X^\pi(S_t)$ that will return a feasible vector $x_t$.  We defer to later the challenge of designing a good policy.
  \item[Exogenous information variables] $W_{t+1} = (\phat_{t+1})$, where $\phat_{t+1}$ is the price charged at time $t+1$ as reported by the grid.  The price data could be from historical data, or field observations (for an online application), or a mathematical model.
  \item[Transition function] $S_t = S^M(S_t,x_t,W_{t+1})$, which consists of the equations:
  \bn
  R_{t+1} &=& R_t + \eta (x^{GB}_t + x^{EB}_t - x^{BD}_t),\label{eq:energytransitionR} \\
  p_{t+1} &=& \phat_{t+1}.\label{eq:energytransitionp}
  \en
  The transition function needs to include an equation for each element of the state variable.  In real applications, the transition function can become quite complex (``500 lines of Matlab code'' was how one professional described it).
  \item[Objective function] Let $C(S_t,x_t)$ be the one-period contribution function given by
  \bns
  C(S_t,x_t) = p_t x_t.
  \ens
  We wish to find a policy $X^\pi(S_t)$ that maximizes profits over time, so we use the cumulative reward objective, giving us
  \bn
  \max_\pi \E_{S_0} \E_{W_1,\ldots,W_T|S_0} \left\{\sum_{t=0}^T C(S_t, X^\pi(S_t))|S_0\right\}, \label{eq:energystorageobjective}
  \en
  where $S_{t+1} = S^M(S_t,x_t=X^\pi(S_t),W_{t+1})$ and where we are given an information process $(S_0, W_1, W_2, \ldots, W_T)$.
\end{description}
Of course, this is a very simple problem.  We are going to return to this problem in section \ref{sec:statevariables} where we will use a series of modifications to illustrate how to model state variables with increasing complexity.

\subsection{Pure learning problem}
\label{sec:purelearning}
An important class of problems are pure learning problems, which are widely studied in the literature under the umbrella of multiarmed bandit problems.  Assume that $x\in\Xcal=\{x_1, \ldots, x_M\}$ represents different configurations for manufacturing a new model of electric vehicle which we are going to evaluate using a simulator.  Let $\mu_x = \E_W F(x,W)$ be the expected performance if we could run an infinitely long simulation.  We assume that a single simulation (of reasonable duration) produces the performance
\bns
\Fhat_x = \mu_x + \varepsilon,
\ens
where $\varepsilon \sim N(0, \sigma^2_W)$ is the noise from running a single simulation.

Assume we use a Bayesian model (we could do the entire exercise with a frequentist model), where our prior on the truth $\mu_x$ is given by $\mu_x \sim N(\mubar^0_x, \sigmabar^{2,0}_x)$.  Assume that we have performed $n$ simulations, and  that $\mu_x \sim N(\mubar^n_x, \sigmabar^{2,n}_x)$.  Our belief $B^n$ about $\mu_x$ after $n$ simulations is then given by
\bn
B^n = (\mubar^n_x,\sigmabar^{2,n}_x)_{x\in\Xcal}. \label{eq:learningstate}
\en
For convenience, we are going to define the {\it precision} of an experiment as $\beta^W = 1/\sigma^2_W$, and the precision of our belief about the performance of drug $x$ as $\beta^n_x = 1/\sigmabar^{2,n}_x$.

If we choose to try drug $x^n$ and then run the $n+1st$ experiment and observe $\Fhat^{n+1} = F(x^n,W^{n+1})$, we update our beliefs using
\bn
\mubar^{n+1}_x & = & \frac{\beta^n_x \mubar^n_x + \beta^W \Fhat^{n+1}_x}{\beta^n_x + \beta^W},    \label{eq:learningtransitionmubar}\\
\beta^{n+1}_x  & = & \beta^n_x + \beta^W,  \label{eq:learningtransitionsigmabar}
\en
if $x=x^n$; otherwise, $\mubar^{n+1}_x = \mubar^n_x$ and $\beta^{n+1}_x = \beta^n_x$.  These updating equations assume that beliefs are independent; it is a minor extension to allow for correlated beliefs.

Also, these equations are for a Bayesian belief model.  In section \ref{sec:timeseriesprice} we are going to illustrate learning with a frequentist belief model.

We are now ready to state our model using the canonical framework:

\begin{description}
  \item[State variables] The state variable is the belief $S^n = B^n$ given by equation \eqref{eq:learningstate}.
  \item[Decision variables] The decision variable is the configuration $x \in \Xcal$ that we wish to test next, which will be determined by a policy $X^\pi(S^n)$.
  \item[Exogenous information] This is the simulated performance given by $\Fhat^{n+1}_{x^n}$.
  \item[Transition function] These are given by equations \eqref{eq:learningtransitionmubar}-\eqref{eq:learningtransitionsigmabar} for updating the beliefs.
  \item[Objective function] This is a state-independent problem (the only state variable is our belief about the performance). We have a budget to run $N$ simulations of different configurations.  When the budget is exhausted, we choose the best design according to
      \bns
      x^{\pi,N} = \argmax_{x\in\Xcal} \mubar^N_x,
      \ens
      where we introduce the policy $\pi$ because $\mubar^N_x$ has been estimated by running experiments using experimentation policy $X^\pi(S^n)$.  The performance of a policy $X^\pi(S^n)$ is given by
      \bns
      F^\pi = \E_{S^0} \E_{W^1,\ldots, W^N|S^0} \E_{\What|S^0} F(x^{\pi,N},\What).
      \ens
      Our goal is to then solve
      \bns
      \max_\pi \E F^\pi(S^0).
      \ens
\end{description}

Note that when we made the transition from an energy storage problem to a learning problem, the modeling framework remained the same.  The biggest change is the state variable, which is now a belief state.

The modeling of learning problems is somewhat ragged in the academic literature.  In a tutorial on reinforcement learning, \cite{lazaric2019} states that bandit problems do not have a state variable (!!).  In contrast, there is a substantial literature on bandit problems in the applied probability community that studies ``Gittins indices'' that is based on solving Bellman's equation exactly where the state is the belief (see \cite{gittins2011} for a nice overview of this field).

Our position is that a belief state is simply part of the state variable, which may include elements that we can observe perfectly, as well as beliefs about parameters that can only be estimated.  This leaves us with the challenge of designing policies.

\section{Designing policies}
\label{sec:designingpolicies}
There are two fundamental strategies for designing policies, each of which can be further divided into two classes, producing four classes of policies:
\begin{description}
  \item[Policy search] - Here we use any of the objective functions  \eqref{eq:stateindependentfinalreward} -  \eqref{eq:statedependentcumulativereward} to search within a family of functions to find the policy that works best.  Policies in the policy-search class can be further divided into two classes:
      \begin{description}
        \item[Policy function approximations (PFAs)] PFAs are analytical functions that map states to actions.  They can be lookup tables (if the chessboard is in this state, then make this move), or linear models which might be of the form
            \bns
            X^{PFA}(S^n|\theta) = \sum_{f\in\Fcal} \theta_f \phi_f(S^n).
            \ens
            PFAs can also be nonlinear models (buy low, sell high is a form of nonlinear model), or even a neural network.
        \item[Cost function approximations (CFAs)] CFAs are parameterized optimization models.  A simple one that is widely used in pure learning problems, called interval estimation, is given by
            \bns
            X^{CFA-IE}(S^n|\theta^{IE}) = \argmax_{x\in\Xcal} (\mubar^n_x + \theta^{IE} \sigmabar^n_x).
            \ens
            The CFA might be a large linear program, such as that used to schedule aircraft where the amount of slack for weather delays is set at the $\theta$-percentile of the distribution of travel times.  We can write this generally as
            \bns
            X^{CFA}(S^n|\theta) = \argmax_{x\in\Xcal^\pi(\theta)} \Cbar^\pi(S^n,x|\theta),
            \ens
            where $\Cbar^\pi(S^n,x|\theta)$ might be a parametrically modified objective function (e.g. with penalties for being late), while $\Xcal^\pi(\theta)$ might be parametrically modified constraints (think of buffer stocks and schedule slack).
      \end{description}
  \item[Lookahead approximations] - We can create an optimal policy if we could solve
      \bn
      X^*_t(S_t) = \argmax_{x_t} \hspace{-.03in}\left(\hspace{-.03in}C(S_t,x_t) + \E \left\{ \left. \max_\pi  \E \left\{\left.\sum_{t'=t+1}^T C(S_{t'},X^\pi_{t'}(S_{t'})) \right| S_{t+1}\right\}\right|S_t,x_t\right\}\right)\hspace{-.05in}. \label{eq:optimalLApolicy}
      \en
      In practice, equation \eqref{eq:optimalLApolicy} cannot be computed, so we have to resort to approximations. There are two approaches for creating these approximations:
      \begin{description}
        \item[Value function approximations (VFAs)] The ideal VFA policy involves solving Bellman's equation
        \bn
        V_t(S_t) = \max_x\big(C(S_t,x) + \E \{V_{t+1}(S_{t+1})|S_t,x\}\big). \label{eq:bellman}
        \en
        We can build a series of policies around Bellman's equation:
        \bn
        X^{VFA}(S_t|\theta) &=& \argmax_{x\in\Xcal_t}\big(C(S_t,x) + \E \{V_{t+1}(S_{t+1})|S_t,x\}\big),\label{eq:vfa0}\\
                            &=& \argmax_{x\in\Xcal_t}\big(C(S_t,x) + \E \{\Vbar_{t+1}(S_{t+1}|\theta)|S_t,x\}\big),\label{eq:vfa1}\\
                            &=& \argmax_{x\in\Xcal_t}\big(C(S_t,x) +  \Vbar^x_t(S^x_t|\theta)\big),\label{eq:vfa2}\\
                            &=& \argmax_{x\in\Xcal_t}\big(C(S_t,x) + \sum_{f\in\Fcal}\theta_f \phi_f(S_t,x)\big),\label{eq:vfa3}\\
                            &=& \argmax_{x\in\Xcal_t} \Qbar(S_t,x|\theta). \label{eq:vfa4}
        \en
        The policy given in equation \eqref{eq:vfa0} would be optimal if we could compute $V_{t+1}(S_{t+1})$ from \eqref{eq:bellman} exactly.  Equation \eqref{eq:vfa1} replaces the value function with an approximation, which assumes that a) we can come up with a reasonable approximation and b) we can compute the expectation.  Equation \eqref{eq:vfa2} eliminates the expectation by using the post-decision state $S^x$ (see \cite{PowellADP2011} for a discussion of post-decision states).  Equation \eqref{eq:vfa3} introduces a linear model for the value function approximation. Finally, equation \eqref{eq:vfa4} writes the equation in the form of $Q$-learning used in the reinforcement learning community.
        \item[Direct lookaheads (DLAs)] The second approach is to create an {\it approximate lookahead model}.  If we are making a decision at time $t$, we represent our lookahead model using the same notation as the base model, but replace the state $S_t$ with $\Stilde_{tt'}$, the decision $x_t$ with $\xtilde_{tt'}$ which is determined with policy $\Xtilde^{\tilde \pi}(\Stilde_{tt'})$, and the exogenous information $W_t$ with $\Wtilde_{tt'}$.  This gives us an approximate lookahead policy

        {\small
        \bn
        \hspace{-0.20in} X^{DLA}_t(S_t) \hspace{-0.10in}&=&\hspace{-0.10in} \argmax_{x_t} \left(C(S_t,x_t) + \Etilde \left\{\max_{{\tilde \pi}} \Etilde \left\{\sum_{t'=t+1}^T C(\Stilde_{tt'},\Xtilde^{\tilde \pi}(\Stilde_{tt'})) | \Stilde_{t,t+1}\right\}|S_t,x_t\right\}\right). \label{eq:optimalpolicyLAapproxpi}
        \en
        }
      \end{description}
\end{description}

We claim that these four classes are universal, which means that any policy designed for any sequential decision problem will fall in one of these four classes, or a hybrid of two or more.  We further insist that all four classes are important. \cite{PowellMeisel2016} demonstrates that each of the four classes of policies may work best, depending on the characteristics of the datasets, for the energy storage problem described in section \ref{sec:energystorage}.  Further, all four classes of policies have been used (by different communities) for pure learning problems.

We emphasize that these are four meta-classes.  Choosing one of the meta-classes does not mean that you are done, but it does help guide the process.  Most (almost all) of the literature on decisions under uncertainty is written with one of the four classes in mind.  We think all four classes are important.  Most important is that at least one of the four classes {\it will} work, which is why we insist on ``{\it model first, then solve}.''

It has been our experience that the most consistent error made in the modeling of sequential decision problems arises with state variables, so we address this next. It is through the state variable that we can model problems with physical states, belief states or both.  Regardless of the makeup of the state variable, we will still turn to the four classes of policies for making decisions.

\section{State variables}
\label{sec:statevariables}
The definition of a state variable is central to the proper modeling of any sequential decision problem, because it captures the information available to make a decision, along with the information needed to compute the objective function and the transition function.
The policy is the function that derives information from the state variable to make decisions.

We are going to start in section \ref{sec:briefhistory} with a brief history of state variables.  In section \ref{sec:moderndefinition} we offer our own definition of a state variable (this is taken from \cite{PowellRLSO2020} which in turn is based on the definition offered in \cite{PowellADP2011}[Chapter 5, available at http://adp.princeton.edu].  Section \ref{sec:moreillustrations} then provides a series of extensions of our energy storage problem to illustrate history-dependent problems, passive and active learning, and the widely overlooked issue of modeling rolling forecasts.  We close by giving a probabilist's measure-theoretic perspective of information and state variables in section \ref{sec:probabilistsperspective}.

\subsection{A brief history of state variables}
\label{sec:briefhistory}
Our experience is that there is an almost universal misunderstanding of what is meant by a ``state variable.''  Not surprisingly, interpretations of the term ``state variable'' vary between communities.  An indication of the confusion can be traced to attempts to define state variables.  For example, Bellman introduces state variables with ``we have a physical system characterized at any stage by a small set of parameters, the {\it state variables}'' \citep{Be57}.  Puterman's now classic text introduces state variables with ``At each decision epoch, the system occupies a {\it state}.'' \citep{Puterman05}[p. 18] (in both cases, the italicized text was included in the original text).  As of this writing, Wikipedia offers ``A state variable is one of the set of variables that are used to describe the mathematical ‘state’ of a dynamical system.''  Note that all three references use the word ``state'' in the definition of state variable (which means it is not a proper definition).

In fact, the vast majority of books that deal with sequential decision problems in some form do not offer a definition of a state variable, with one notable exception: the optimal control community.  There, we have found that books in optimal control routinely offer an explicit definition of a state variable.  For example, \cite{Ki98} offers:
\begin{changemargin}{1.0cm}{1.0cm}
{\it
A state variable is a set of quantities $x_1(t), x_2(t), \ldots$ [WBP: the controls community uses $x(t)$ for the state variable] which if known at time $t=t_0$ are determined for $t \geq t_0$ by specifying the inputs for $t \geq t_0$.
}
\end{changemargin}
\cite{CaLa2008} has the definition:
\begin{changemargin}{1.0cm}{1.0cm}
{\it
The  state of a system at time $t_0$ is the information required at $t_0$ such that the output [cost] $y(t)$ for all $t\geq t_0$ is uniquely determined from this information and from [the control] $u(t), ~t\geq t_0$.
}
\end{changemargin}
We have observed that the pattern of designing state variables is consistent across books in deterministic control, but not stochastic control.  We feel this is because deterministic control books are written by engineers who need to model real problems, while stochastic control books are written by mathematicians.

There is a surprisingly widespread belief that a system can be ``non-Markovian'' but can be made ``Markovian'' by adding to the state variable.  This is nicely illustrated in \cite{cinlar2011}:
\begin{changemargin}{1.0cm}{1.0cm}
{\it
The definitions of ``time'' and ``state'' depend on the application at hand and the demands of mathematical tractability.  Otherwise, if such practical considerations are ignored, every stochastic process can be made Markovian by enhancing its state space sufficiently.
}
\end{changemargin}

We agree with the basic principle expressed in the controls books, which can all be re-stated as saying ``A state variable is all the information you need (along with exogenous inputs) to model the system from time $t$ onward.''  Our only complaint is that this is a bit vague.

On the other hand, we disagree with the widely held belief that stochastic systems ``can be made Markovian'' which runs against the core principle in the definitions in the optimal control books that the state variable is all the information needed to model the system from time $t$ onward.  If it is {\it all} the information, then it is Markovian by construction.

There are two key areas of misunderstanding that we are going to address with our discussion.  The first is a surprisingly widespread misunderstanding about ``Markov'' vs. ``history-dependent'' systems.  The second, and far more subtle, arises when there are hidden or unobservable variables.

\subsection{A modern definition}
\label{sec:moderndefinition}
We offer two definitions depending on whether we have a system where the structure of the policy has been specified, and when it has not (this is taken from \cite{PowellRLSO2020}).
\begin{changemargin}{1.0cm}{1.0cm}
{\small
A {\bf state variable} is:
\begin{description}
\item[a) Policy-dependent version] A function of history that, combined with the exogenous information (and a policy), is necessary and sufficient to compute the decision function (the policy), the cost/contribution function,  and the transition function.
\item[b) Optimization version] A function of history that, combined with the exogenous information, is necessary and sufficient to compute the cost or contribution function, the constraints, and the transition function.
\end{description}
}
\end{changemargin}
Both of these definitions lead us to our first claim:
\begin{changemargin}{1.0cm}{1.0cm}
{\it
Claim 1: All properly modeled systems are Markovian.
}
\end{changemargin}
But stay tuned; later, we are going to argue the opposite, but there will be a slight change in the wording that explains the apparent contradiction.

Note that both of these definitions are consistent with those used in the controls community, with the only difference that we have specified that we can identify state variables by looking at the requirements of three functions: the cost/contribution function, the constraints (which is a form of function), and the transition function.

One issue that we are going to address arises when we ask ``What is a transition function?''

\begin{changemargin}{1.0cm}{1.0cm}
{\it
The {\bf transition function}, which we write $S_{t+1} = S^M(S_t,x_t,W_{t+1})$, is the set of equations that describes how each element of the state variable $S_t$ evolves over time.
}
\end{changemargin}
We quickly see that we have circular reasoning: a state variable includes the information we need to model the transition function, and the transition function is the equations that describe the evolution of the state variables.  It turns out that this circular logic is unavoidable, as we illustrate later (in section \ref{sec:rollingforecast}).

\begin{figure}[tb]
\begin{center}
    \includegraphics[width=4.0in]{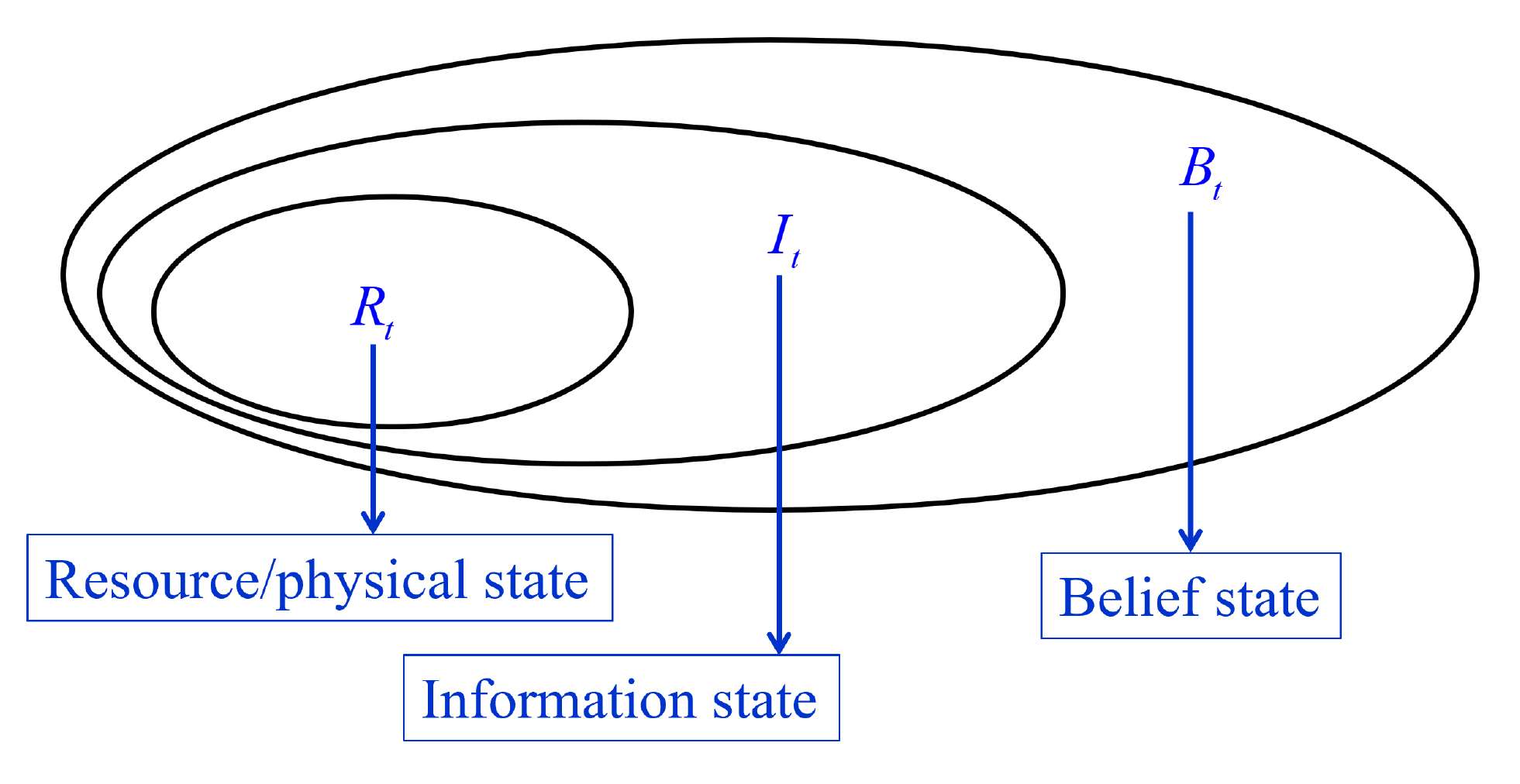}
    \caption{Physical state variables $R_t$, as a subset of other information $I_t$, as a subset of belief state variables $B_t$.}
    \label{fig:nestedstatevariables}
\end{center}
\end{figure}

We have found it useful to identify three types of state variables:
\begin{description}
  \item[Physical state $R_t$] - The physical state captures inventories, the location of a device on a graph, the demand for a product, the amount of energy available from a wind farm, or the status of a machine.  Physical states typically appear in the right hand sides of constraints.
  \item[Other information $I_t$] - The ``other information'' variable is literally any other information about observable parameters not included in $R_t$.
  \item[Belief state $B_t$] - The belief state $B_t$ captures the parameters of a probability distribution describing unobservable parameters.  This could be the mean and variance of a normal distribution, or a set of probabilities.
\end{description}
We present the three types of state variables as being in distinct classes, but it is more accurate to describe them as a series of nested sets as depicted in figure \ref{fig:nestedstatevariables}.  The physical state variables describe quantities that constrain the system (inventories, location of a truck, demands) that are known perfectly.  We then describe $I_t$ as ``other information'' but it might help to think of $I_t$ as any parameter that we observe perfectly, which could include $R_t$.  Finally, $B_t$ is the parameters of probability distributions for any quantity that we do not know perfectly, but a special case of a probability distribution is a point estimate with zero variance, which could include $I_t$ (and then $R_t$).  Our choice of the three variables is designed purely to help with the modeling process.

We explicitly model the ``resource state'' because we have found in some communities (and this is certainly true of operations research) that people tend to equate ``state'' and ``physical state.'' We do not offer an explicit definition of $R_t$, although we note that it typically includes dynamic information in right-hand side constraints.  $R_{ti}$ might be the number of units of blood of type $i$ at time $t$; $R_{ta}$ might be the number of resources with attribute vector $a$.  For example, $R_{ta}$ could be how much we have invested in an asset, where $a$ captures the type of asset, how long it has been invested, and other information (such as the current price of the asset).

Some authors find it convenient to distinguish between two types of states:
\begin{description}
  \item[Exogenous states] These are dynamically varying parameters that evolve purely from an exogenous process.
  \item[Controllable states] These are the state variables that are directly or indirectly affected by decisions.
\end{description}
In the massive class of problems known as ``dynamic resource allocation,'' $R_t$ would be the physical state, and this would also be the controllable state.  However, there may be variables in $I_t$ that are also controllable (at least indirectly).  There will also be states (such as water in a reservoir, or the state of disease in a patient) that evolve due to a mixture of exogenous and controllable processes.

Later we are going to illustrate the widespread confusion in the handling of ``states'' (physical states in our language) and ``belief states'' in the literature on partially observable Markov decision processes (POMDPs).

\subsection{More illustrations}
\label{sec:moreillustrations}
We are going to use our energy storage problem to illustrate the handling of so-called ``history-dependent'' problems (in section \ref{sec:timeseriesprice}), followed by examples of passive and active learning (in sections \ref{sec:passivelearning} and \ref{sec:activelearning}), closing with an illustration of the circular logic for defining state variables and transition functions using rolling forecasts (in section \ref{sec:rollingforecast}).  The material in this section is taken from \cite{PowellRLSO2020}.

\subsubsection{With a time-series price model}
\label{sec:timeseriesprice}
Our basic model assumed that prices evolved according to a purely exogenous process (see equation \eqref{eq:energytransitionp}).  Now assume that it is governed by the time series model
\bn
p_{t+1} &=& \theta_0 p_t + \theta_1 p_{t-1} + \theta_2 p_{t-2} + \varepsilon_{t+1}. \label{eq:priceprocesswithoutlearning}
\en
A common mistake is to say that $p_t$ is the ``state'' of the price process, and then observe that it is no longer Markovian (it would be called ``history dependent''), but ``it can be made Markovian by expanding the state variable,'' which would be done by including $p_{t-1}$ and $p_{t-2}$.  According to our definition of a state variable, the state is all the information needed to model the process from time $t$ onward, which means that the state of our price process is $(p_t, p_{t-1}, p_{t-2})$.  This means our system state variable is now
\bns
S_t = \big(R_t, (p_t,p_{t-1},p_{t-2})\big)).
\ens
We then have to modify our transition function so that the ``price state variable'' at time $t+1$ becomes $(p_{t+1}, p_t, p_{t-1})$.

\subsubsection{With passive learning}
\label{sec:passivelearning}
We implicitly assumed that our price process in equation \eqref{eq:priceprocesswithoutlearning} was governed by a model where the coefficients $\theta = (\theta_0, \theta_1, \theta_2)$ were known.  Now assume that the vector $\theta$ is unknown, which means we have to use estimates $\thetabar_t = (\thetabar_{t0}, \thetabar_{t1}, \thetabar_{t2})$, which gives us the price model
\bn
p_{t+1} &=& \thetabar_{t0} p_t + \thetabar_{t1} p_{t-1} + \thetabar_{t2} p_{t-2} + \varepsilon_{t+1}. \label{eq:priceprocesspassivelearning}
\en
We have to adaptively update our estimate $\thetabar_t$ which we can do using recursive least squares.  To do this, let
\bns
\pbar_t                      &=& (p_t, p_{t-2}, p_{t-2})^T,\\
\Fbar_t(\pbar_t|\thetabar_t) &=& (\pbar_t)^T \thetabar_t.
\ens
We perform the updating using a standard set of updating equations given by
\bn
\thetabar_{t+1} &=& \thetabar_t + \frac{1}{\gamma_t} M_t \pbar_t \varepsilon_{t+1}, \label{eq:learning1} \\
\varepsilon_{t+1} &=& \Fbar_t(\pbar_t|\thetabar_t) - p_{t+1}, \label{eq:learning2} \\
M_{t+1} &=& M_t - \frac{1}{\gamma_t} M_t (\pbar_t) (\pbar_t)^T M_t, \label{eq:learning3} \\
\gamma_t &=& 1- (\pbar_t)^T M_t \pbar_t. \label{eq:learning4}
\en
To compute these equations, we need the three-element vector $\thetabar_t$ and the $3 \times 3$ matrix $M_t$.  These then need to be added to our state variable, giving us
\bns
S_t = \big(R_t, (p_t,p_{t-1},p_{t-2}), (\thetabar_t,M_t)\big).
\ens
We then have to include equations \eqref{eq:learning1} - \eqref{eq:learning4} in our transition function.

\subsubsection{With active learning}
\label{sec:activelearning}
We can further generalize our model by assuming that our decision $x_t$ to buy or sell energy from or to the grid can have an impact on prices.  We might propose a modified price model given by
\bn
p_{t+1} &=& \thetabar_{t0} p_t + \thetabar_{t1} p_{t-1} + \thetabar_{t2} p_{t-2} + \thetabar_{t3} x_t + \varepsilon_{t+1}. \label{eq:priceprocessactivelearning}
\en
All we have done is introduce a single term $\thetabar_{t3} x_t$ (which specifies how much we buy/sell from/to the grid) to our price model.  Assuming that $\theta_3 >0$, this model implies that purchasing power from the grid ($x_t > 0$) will increase grid prices, while selling power back to the grid ($x_t <0$) decreases prices.  This means that purchasing a lot of power from the grid (for example) means we are more likely to observe higher prices, which may assist the process of learning $\theta$.  When decisions control or influence what we observe, then this is an example of {\it active learning}, which we saw in section \ref{sec:purelearning} when we described a pure learning problem.

This change in our price model does not affect the state variable from the previous model, aside from adding one more element to $\thetabar_t$, with the required changes to the matrix $M_t$.  The change will, however, have an impact on the policy.  It is easier to learn $\theta$ if there is a nice spread in the prices, which is enhanced by varying $x_t$ over a wide range.  This means trying values of $x_t$ that do not appear to be optimal given our current estimate of the vector $\thetabar_t$.  Making decisions partly just to learn (to make better decisions in the future) is the essence of {\it active learning}, best known in the field of multiarmed bandit problems.

\subsubsection{With rolling forecasts}
\label{sec:rollingforecast}
We are going to assume that we are given a rolling forecast from an outside source.  This is quite common, and yet is surprisingly overlooked in the modeling of dynamic systems (including inventory/storage systems, for which there is an extensive literature).  We are going to use rolling forecasts to illustrate the interaction between the modeling of state variables and the creation of the transition function.

Imagine that we are modeling the energy $E_t$ from wind, which means we would have to add $E_t$ to our state variable.  We need to model how $E_t$ evolves over time.  Assume we have a forecast $f^E_{t,t+1}$ of the energy $E_{t+1}$ from wind, which means
\bn
E_{t+1} = f^E_{t,t+1} + \varepsilon_{t+1,1},  \label{eq:energytransitionE1}
\en
where $\varepsilon_{t+1,1} \sim N(0, \sigma^2_\varepsilon)$ is the random variable capturing the one-period-ahead error in the forecast.

Equation \eqref{eq:energytransitionE1} needs to be added to the transition equations for our model.  However, it introduces a new variable, the forecast $f^E_{t,t+1}$, which must now be added to the state variable.  This means we now need a transition equation to describe how $f^E_{t,t+1}$ evolves over time.  We do this by using a two-period-ahead forecast, $f^E_{t,t+2}$, which is basically a forecast of $f^E_{t+1,t+2}$, plus an error, giving us
\bn
f^E_{t+1,t+2} = f^E_{t,t+2} + \varepsilon_{t+1,2},  \label{eq:energytransitionE2}
\en
where $\varepsilon_{t+1,2} \sim N(0, 2\sigma^2_\varepsilon)$ is the two-period-ahead error (we are assuming that the variance in a forecast increases linearly with time).  Now we have to put $f^E_{t,t+2}$ in the state variable, which generates a new transition equation.   This generalizes to
\bn
f^E_{t+1,t'} = f^E_{t,t'} + \varepsilon_{t+1,t'-t},  \label{eq:energytransitionEt}
\en
where $\varepsilon_{t+1,t'-t} \sim N(0, (t'-t)\sigma^2_\varepsilon)$.  This process illustrates the back and forth between defining the state variable and creating the transition function that we hinted at earlier.

This stops, of course, when we hit the planning horizon $H$.  This means that we now have to add
\bns
f^E_t = (f^E_{tt'})_{t'=t+1}^{t+H}
\ens
to the state variable, with the transition equations \eqref{eq:energytransitionEt} for $t'=t+1, \ldots, t+H$.  Combined with the learning statistics, our state variable is now
\bns
S_t = \big((R_t, E_t), (p_t,p_{t-1},p_{t-2}), (\thetabar_t,M_t),f^E_t\big).
\ens

It is useful to note that we have a nice illustration of the three elements of our state variable:
\bns
(R_t, E_t) &=& \textwrap{The physical state variables,}\\
(p_t, p_{t-1}, p_{t-2}) &=& \textwrap{other information,}\\
((\thetabar_t,M_t),f^E_t) &=& \textwrap{the belief state, since these parameters determine the distribution of belief about variables that are not known perfectly.}
\ens

\subsection{A probabilist's perspective of information}
\label{sec:probabilistsperspective}
We would be remiss in a discussion of state variables if we did not cover how the mathematical probability community thinks of ``information'' and ``state variables.''  We note that this section is completely optional, as will be seen by the end of the section.

We begin by introducing what is widely known as boilerplate language when describing stochastic processes.

\begin{changemargin}{1.0cm}{1.0cm}
{\it
Let $(S_0, W_1, W_2, \ldots, W_T)$ be the sequence of exogenous information variables, beginning with the initial state (that may contain a Bayesian prior), followed by the exogenous information contained in $W_t$.  Let $\omega\in\Omega$ be a sample sequence of a truth (contained in $S_0$), and a realization of $W_1, \ldots, W_T$.  Let $\Fcal$ be the $\sigma$-algebra (also written ``sigma-algebra'') on $\Omega$, which captures all the events that might be defined on $\Omega$.  The set $\Fcal$ is the set of all countable unions and complements of the elements of $\Omega$, which is to say every possible event.  Let $\Pcal$ be a probability measure on $(\Omega,\Fcal)$ (if $\Omega$ is discrete, $\Pcal$ would be a probability mass function).  Now let $\Fcal_t = \sigma(S_0,W_1, \ldots, W_t)$ be the $\sigma$-algebra generated by the process $(S_0, W_1, \ldots, W_t)$, which means it reflects the subsets of $\Omega$ that we can identify using the information that has been revealed up through time $t$.  The sequence $\Fcal_0, \Fcal_1, \ldots, \Fcal_t$ is referred to as a filtration, which means that $\Fcal_t \subseteq \Fcal_{t+1}$ (as more information is revealed, we are able to see more fine-grained events on $\Omega$, which acts like a sequence of filters with increasingly finer openings).
}
\end{changemargin}
This terminology is known as boilerplate because it can be copied and pasted into any model with a stochastic process $(S_0, W_1, \ldots, W_T)$, and does not change with applications (readers are given permission to copy this paragraph word for word, but as we show below, it is not necessary).

We need this language to handle the following issue.  In a deterministic problem, the decisions $x_0, x_1, \ldots, x_T$ represent a sequence of numbers (these can be scalars or vectors).  In a stochastic problem, there is a decision $x_t$ for each sample path $\omega$ which represents a realization of the entire sequence $(S_0, W_1, \ldots, W_t, \ldots, W_T)$.  This means that if we write $x_t(\omega)$, it is as if we are telling $x_t$ the {\it entire} sample path, which means it gets to see what is going to happen in the future.

We fix this by insisting that the function $x_t(\omega)$ be ``$\Fcal_t$-measurable,'' which means that $x_t$ is not allowed to depend on the outcomes of $W_{t+1}, \ldots, W_T$.  We get the same behavior if we write $x_t$ explicitly as a function (that we call a policy) $X^\pi(S_t)$ that depends on just the information in the state $S_t$.  Note that the state $S_t$ is purely a function of the history $S_0, W_1, \ldots, W_t$, in addition to the decisions $x_0  = X^\pi(S_0), x_1 = X^\pi(S_1), \ldots, x_t = X^\pi(S_t)$.

Theoreticians will use $\Fcal_t$ or $S_t$ to represent ``information,''  but they are not equivalent.  $\Fcal_t$ contains all the information in the exogenous sequence $(S_0, W_1, \ldots, W_t)$.  The state $S_t$, on the other hand, is constructed from the sequence $(S_0, W_1, \ldots, W_t)$, but only includes the information we need to compute the objective function, constraints and transition function.  Also, $S_t$ can always be represented as a vector of real-valued numbers, while $\Fcal_t$ is a set of events which contain the information needed to compute $S_t$.  The set $\Fcal_t$ is more general, hence its appeal to mathematicians, while $S_t$ contains the information we actually need to model our problem.

The state $S_t$ can be viewed from three different perspectives, depending on what time it is:
\begin{itemize}
\item[1)] We are at time $t$ - In a sequential decision problem, if we are talking about $S_t$ then it usually means we are at time $t$, in which case $S_t$ is a particular realization of a set of numbers that capture everything we need from history to model our system moving forward.
\item[2)] We are at time $t=0$ - We might be trying to choose the best policy (or some other fixed parameter), in which case we are at time 0, and $S_t$ would be a random variable since we do not know what state we will be in at time $t$ when we are at time $t=0$.
\item[3)] We are at time $t=T$ - Finally, a probabilist sits at time $t=T$ and sees all the outcomes $\Omega$ (and therefore all the events in $\Fcal$), but from his perspective at time $T$, if you ask him a question about $S_t$ at time $t$, he will recognize only events in $\Fcal_t$ (remember that each event in $\Fcal_t$ is a subset of sample paths $\omega\in \Omega$).  For example, if we are running simulations using historical data, and we cheat and use information from time $t' > t$ to make a decision at time $t$, that implies we are seeing an event that is in $\Fcal_{t'}$, but which is not in $\Fcal_t$.  In such a case, our decision would not be ``$\Fcal_t$-measurable.''
\end{itemize}

Readers without training in measure-theoretic probability will find this language unfamiliar, even threatening.  We will just note that the following statements are all completely equivalent.
\begin{itemize}
  \item[1)] The policy $X^\pi(S_t)$ (or decision $x_t)$ is $\Fcal_t$-measurable.
  \item[2)] The policy $X^\pi(S_t)$ (or decision $x_t)$ is nonanticipative.
  \item[3)] The policy $X^\pi(S_t)$ (or decision $x_t)$ is ``adapted.''
  \item[4)] The policy $X^\pi(S_t)$ (or decision $x_t$) is a function of the state $S_t$.
\end{itemize}
Readers without formal training in measure-theoretic probability will likely find statement (4) to be straightforward and easy to understand.  We are here to say that you only need to understand statement (4), which means you can write models (and even publish papers) without any of the other formalism in this section.

\section{Partially observable Markov decision processes}
\label{sec:POMDP}
Partially observable Markov decision processes (POMDPs) broadly describe any sequential decision problem that involves learning an environment that cannot be precisely observed.  However, it is most often associated with problems where decisions can affect the environment, which was not the case in our pure learning problems in section \ref{sec:purelearning}.

We are going to describe our environment in terms of a set of parameters that we are trying to learn.  It is helpful to identify three classes of problems:
\begin{description}
  \item[1) Static unobservable parameters] - These are problems where we are trying to learn the values of a set of static parameters, which might be the response of a function given different inputs, or the parameters characterizing an unknown function.  These experiments could be run in a simulator or laboratory, or in the field. Examples are:
  \begin{itemize}
    \item The strength of material resulting from the use of different catalysts.
    \item Designing a business system using a simulator.  We might be designing the layout of an assembly line, evaluating the number of aircraft in a fleet, or finding the best locations of warehouses in a logistics network.
    \item Evaluating the parameters of a policy for stocking inventory or buying stock.
    \item Controlling robots moving in a static but unknown environment.
  \end{itemize}
  \item[2) Dynamic unobservable parameters] - Now we are trying to learn the value of parameters that are evolving over time.  These come in two flavors:
  \begin{description}
    \item[2a) Exogenous, uncontrollable process] - These are problems where environmental parameters will evolve over time due to a purely exogenous source:
    \begin{itemize}
       \item Demand for hotel rooms as a function of price in changing market conditions (where our decisions do not affect the market).
       \item Robots moving in an uncertain environment that is changing due to weather.
       \item Finding the best path through a congested network after a major road has been closed due to construction forcing people to explore new routes.
       \item Finding the best price for a ridesharing fleet to balance drivers and riders.  The best price evolves as the number of drivers and riders changes over the course of the day.
    \end{itemize}
    \item[2b) Controllable process] - These are problems where the controlling agent makes decisions that directly or indirectly affect environmental parameters:
    \begin{itemize}
       \item Equipment maintenance - We perform inspections to determine the state of the machine, and then perform repairs which changes the state of the equipment.
       \item Medical treatments - We test for a disease, and then treat using drugs which changes the progression of the disease.
       \item Managing a utility truck after a storm, where the truck is both observing and repairing damaged lines, while we control the truck.
       \item Invasive plant or animal species management - We perform inspections, and implement steps to mitigate further spread of the invasive species.
        \item Spread of the flu - We can take samples from the population, and then administer flu vaccinations to reduce incidence of the disease.
    \end{itemize}
  \end{description}
\end{description}

Class (1) represents our pure learning problems (which we first touched on in section \ref{sec:purelearning}), often referred to as multiarmed bandits, although the arms (choices) may be continuous and/or vector valued.  The important characteristic of class (1) is that our underlying problem (the environment) is assumed to be static.  Also, we may be learning in the field, where we are interested in optimizing cumulative rewards (this is the classic bandit setting) or in a laboratory, where we are only interested in the final performance.

Class (2) covers problems where the environment is changing over time.  Class (2a) covers problems where the environment is evolving exogenously.  This has been widely studied under the umbrella of ``restless bandits.''  The modeling of these systems is similar to class (1).  Class (2b) arises when our decisions directly or indirectly affect the parameters that cannot be observed, which is the domain of POMDPs. We are going to illustrate this with a problem to treat flu in a population.

Earlier we noted that Markov decision problems are often represented by the tuple $(\Scal, \Xcal, P, r)$ (we use $\Xcal$ for action space, but standard notation in this community is to use $a\in\Acal$ for action).  The POMDP community extends this representation by representing the POMDP as the tuple $(\Scal, \Xcal, P, r, \Wcal^{obs}, P^{obs})$ where $\Scal$, $\Xcal$, $P$ and $r$ are as they were with the basic MDP, $\Wcal^{obs}$ is the space of possible observations that can be made of the environment, and $P^{obs}$ is the ``observation function'' where
\bns
P^{obs}(w^{obs}|s) &=& Prob[W^{obs}=w^{obs}|s] \\
             &=& \textwrapsmall{the probability we observe outcome $W^{obs}=w^{obs}$ when the unobservable state $S=s$.}
\ens
The notation for modeling POMDPs is not standard.  Different authors may use $Z$ or $Y$ for the observation, and may use $\Zcal$, $\Ycal$ or $\Omega$ for the space of outcomes. Some will use $O$ for outcome and $\Ocal$ for the ``observation function.''  Our choice of $P^{obs}(w^{obs}|s)$ (which is not standard) helps us to avoid the use of ``$O$'' for notation, and makes it clear that it is the probability of making an observation, which parallels our one-step transition matrix $P$ which describes the evolution of the ``state'' $s$.

\remark There is a bit of confusion in the modeling of uncertainty in POMDPs. The tuple $(\Scal, \Xcal, P, r, \Wcal^{obs}, P^{obs})$ represents uncertainty in both the transition matrix $P$, and then through the pair $(\Wcal^{obs}, P^{obs})$ which captures both observations and the probability of making an observation.  Recall that we represent the transition {\it function} in equation \eqref{eq:transition} using $s'=S^M(s,x,w)$ where $W=w$ is our ``exogenous information.''  This is the exogenous information (that is, the random inputs) that drives the evolution of our ``physical system'' with state $S=s$.  We use this random variable to compute our one-step transition matrix from the transition function using
\bn
p(s'|s,x) = \E_W \{\mathbbm{1}_{\{s'=S^M(s,x,w)\}}\}. \label{eq:transitionexpectation}
\en
The random variables $W$ and $W^{obs}$ may be the same.  For example, we may have a queueing system where $W$ is the random number of customers arriving, which we are allowed to observe.  The unknown parameter $s$ might be the arrival rate of customers, which we can estimate using $W$.  In other settings $W$ and $W^{obs}$ may be completely different.  For example, $W$ might be the random transmission of disease in a population, while $W^{obs}$ is the outcome of random samples.

What is important is that the standard modeling representation for POMDPs is as a single, extended problem.  The POMDP framework ensures that the choice of action $a$ (decision $x$ in our notation) is not allowed to see the state of the system, but it does assume that the transition function $P$ and the observation function $P^{obs}$ are both known.  Later we are going to offer a different approach for modeling POMDPs.  Before we do this, it is going to help to have an actual example in mind.

\section{A learning problem: protecting against the flu}
\label{sec:learningflu}
We are going to use the problem of protecting a population against the flu as an illustrative example.  It will start as a learning problem with an unknown but controllable parameter, which is the prevalence of the flu in the population.  We will use this to illustrate different classes of policies, after which we will propose several extensions.

\subsection{A static model}
Let $\mu$ be the prevalence of the flu in the population (that is, the fraction of the population that has come down with the flu).  In a static problem where we have an unknown parameter $\mu$, we make observations using
\bn
W_{t+1} = \mu + \varepsilon_{t+1}, \label{eq:flustaticobservation}
\en
where the noise $\varepsilon_{t+1}\sim N(0,\sigma^2_W)$ is what keeps us from observing $\mu$ perfectly.

We express our belief about $\mu$ by assuming that $\mu \sim N(\mubar_t,\sigmabar^{2}_{t})$.  Since we fix the assumption of normality, we express our belief about $\mu$ as $B_t = (\mubar_t,\sigmabar^{2}_{t})$.  We are again going to express uncertainty using $\beta_t = 1/\sigmabar^{2}_{t}$ which is the precision of our estimate of $\mu$, and $\beta^W = 1/\sigma^2_W$ is the precision of our observation noise $\varepsilon_{t+1}$.

We need to estimate the number of people with the disease by running tests, which produces the noisy estimate $W_{t+1}$.  We represent the decision to run a test by the decision variable $x^{obs}_t$ where
\bns
x^{obs}_t = \left\{\begin{tabular}{cl} 1 & \mbox{if we observe the process and obtain $W_{t+1}$,} \\
                                 0 & \mbox{if no observation is made.}  \end{tabular}\right.
\ens
If $x^{obs}_t = 1$, then we observe $W_{t+1}$ which we can use to update our belief about $\mu$ using
\bn
\mubar_{t+1} & = & \frac{\beta_t \mubar_{t} + \beta^W W_{t+1}}{\beta_t + \beta^W},    \label{eq:learningtransitionmubarflustatic}\\
\beta_{t+1}  & = & \beta_t + \beta^W.  \label{eq:learningtransitionsigmabarflustatic}
\en
If $x^{obs}_t = 0$, then $\mubar_{t+1} = \mubar_t$ and $\beta_{t+1} = \beta_t$.

For this problem our state variable is our belief about $\mu$, which we write
\bns
S_t = B_t = (\mubar_t, \beta_t).
\ens

If this was our problem, it would be an instance of a one-armed bandit.  We might assess a cost for making an observation, along with a cost of uncertainty.  For example, assume we have the following costs:
\bns
c^{obs} &=& \textwrap{The cost of sampling the population to estimate the number of people infected with the flu,}\\
C^{unc}(S_t) &=& \textwrap{the cost of uncertainty,}\\
                  &=& c^{unc} \sigmabar_t,\\
C(S_t,x_t)        &=& c^{obs} x^{obs}_t + C^{unc}(S_t).
\ens
Using this information, we can put this model in our canonical framework as follows:
\begin{description}
  \item[State variables] $S_t = (\mubar_t, \beta_t)$.
  \item[Decision variables] $x_t = x^{obs}_t$ determined by our policy $X^{obs}(S_t)$ (to be determined later).
  \item[Exogenous information] $W_{t+1}$ which is our noisy estimate of how many people have the flu from equation \eqref{eq:flustaticobservation} (and we only obtain this if $x^{obs} = 1$).
  \item[Transition function] Equations \eqref{eq:learningtransitionmubarflustatic} and \eqref{eq:learningtransitionsigmabarflustatic}.
  \item[Objective function] We would write our objective as
      \bn
      \max_\pi \E\left\{\sum_{t=0}^T C(S_t,x_t) | S_0\right\}. \label{eq:objectivestaticflu}
      \en
\end{description}
We now need a policy $X^{obs}(S_t)$ to determine $x^{obs}_t$.  We can use any of the four classes of policies described in section \ref{sec:designingpolicies}.  We sketch examples of policies in section \ref{sec:designingpolicyflu} below.

\subsection{Variations of our flu model}
We are going to present a series of variations of our flu model to bring out different modeling issues:
\label{sec:fluvariations}
\begin{itemize}
\item A time-varying model
\item A time-varying model with drift
\item A dynamic model with a controllable truth
\item A flu model with a resource constraint and exogenous state
\item A spatial model
\end{itemize}
These variations are designed to bring out the modeling issues that arise when we have an evolving truth (with known dynamics), an evolving truth with unknown dynamics (the drift), an unknown truth that we can control (or influence), followed by problems that introduce the dimension of having a known and controllable physical state.

\subsubsection{A time-varying model}
If the true prevalence of the flu is evolving exogenously (as we would expect in this application), then we would write the true parameter as depending on time, $\mu_t$, which might evolve according to
\bn
\mu_{t+1} = \max\{0,\mu_t + \varepsilon^\mu_{t+1}\}, \label{eq:timevaryingmodel}
\en
where $\varepsilon^\mu_{t+1} \sim N(0, \sigma^{\mu,2})$ describes how our truth is evolving.   If the truth evolves with zero mean and known variance $\sigma^{\varepsilon,2}$, our belief state is the same as it was with a static truth (that is, $S_t = (\mubar_t, \beta_t)$).  What does change is the transition function which now has to reflect both the noise of an observation $\varepsilon_{t+1}$ as well as the uncertainty in the evolution of the truth, captured by $\varepsilon^\mu_{t+1}$.

\remark When $\mu$ was a constant, we did not have a problem referring to it as a parameter, whereas the state of our system is the belief which evolves over time (state variables should only include information that changes over time).  When $\mu$ is changing over time, in which case we write it as $\mu_t$, then it is more natural to think of the value of $\mu_t$ as the state of the system, but not observable to the controller.  For this reason, many authors would refer to $\mu_t$ as a {\it hidden state}.  However, we still have the belief about $\mu_t$, which creates some confusion: What is the state variable?  We are going to resolve this confusion below.

\subsubsection{A time-varying model with drift}
Now assume that
\bns
\varepsilon^\mu_{t+1} \sim N(\delta,\sigma^{\varepsilon,2}).
\ens
If $\delta \ne 0$, then it means that $\mu_t$ is drifting higher or lower (for the moment, we are going to assume that $\delta$ is a constant).  We do not know $\delta$, so we would assign a belief such as
\bns
\delta \sim N(\deltabar_t, \sigmabar^{\delta,2}_t).
\ens
Again let the precision be given by $\beta^\delta_t = 1/\sigmabar^{\delta,2}_t$.

We might update our estimate of our belief about $\delta$ using
\bns
{\deltahat}_{t+1} = W_{t+1} - W_t.
\ens
Now we can update our estimate of the mean and variance of our belief about $\delta$ using
\bn
\deltabar_{t+1} & = & \frac{\beta^\delta_t \deltabar_{t} + \beta^W \deltahat_{t+1}}{\beta^\delta_t + \beta^W},    \label{eq:learningtransitiondeltabarflu}\\
\beta^\delta_{t+1}     & = & \beta^\delta_t + \beta^W.  \label{eq:learningtransitiondeltasigmabarflu}
\en
In this case, our state variable becomes
\bns
S_t = B_t = \big((\mubar_t, \beta_t), (\deltabar_t,\beta^\delta_t)\big).
\ens
Here, we are modeling only the belief about $\mu_t$, while $\mu_t$ itself is just a dynamically varying parameter.  This changes in the next example.

\subsubsection{A dynamic model with a controllable truth}
Now consider what happens when our decisions might actually change the truth $\mu_t$.  Let
\bns
x^{vac}_t = \textwrap{the number of vaccination shots we administer in the region.}
\ens
We assume that the vaccination shots reduce the presence of the disease by $\theta^{vac}$ for each vaccinated patient, which is $x^{vac}_t$.  We are going to assume that the decision made at time $t$ is not implemented until time $t+1$.  This gives us the following equation for the truth
\bn
\mu_{t+1} = \max\{0,\mu_t - \theta^{vac}x^{vac}_{t-1} + \varepsilon^\mu_{t+1}\}. \label{eq:flucontrollabletruth}
\en

We express our belief about the presence of the disease by assuming that it is Gaussian where $\mu_t \sim N(\mubar_t, \sigma^2_t)$.  Again letting the precision be $\beta_t = 1/\sigma^2_t$, our belief state is $B_t = (\mubar_t, \beta_t)$, with transition equations similar to those given in equations \eqref{eq:learningtransitionmubar} and \eqref{eq:learningtransitionsigmabar} but adjusted by our belief about what our decision is doing.  If we make an observation (that is, if $x^{obs}_t = 1$), then
\bn
\mubar_{t+1} & = & \frac{\beta_t (\mubar_{t}-\theta^{vac}x^{vac}_{t-1}) + \beta^W W_{t+1}}{\beta_t + \beta^W},    \label{eq:learningtransitionmubarflu}\\
\beta_{t+1}  & = & \beta_t + \beta^W.  \label{eq:learningtransitionsigmabarflu}
\en
If $x^{obs}_t = 0$, then $\mubar_{t+1} = \mubar_t - \theta^{vac}x^{vac}_{t-1}$, and $\beta_{t+1} = \beta_t$.


This setting introduces a modeling challenge: Is the state $\mu_t$?  Or is it the belief $(\mubar_t,\beta_t)$?  When $\mu_t$ was static or evolved exogenously, it seemed clear that the state was our belief about $\mu_t$.  However, now that we can control $\mu_t$, it seems more natural to view $\mu_t$ as the state.  This problem is an instance of a partially observable Markov decision problem.  Later we are going to review how the POMDP community models these problems, and offer a different approach.  

This problem has an unobservable state that is controllable.  The next two problems will introduce the dimension of combining both observable and unobservable states that are both controllable.

\subsubsection{A flu model with a resource constraint and exogenous state}
Now imagine that we have a limited number of vaccinations that we can administer.  Let $R_0$ be the number of vaccinations we have available.  Our vaccinations $x^{vac}_t$ have to be drawn from this inventory.  We might also introduce a decision $x^{inv}_t$ to add to our inventory (at a cost).  This means our inventory evolves according to
\bns
R_{t+1} = R_t + x^{inv}_{t-1} - x^{vac}_{t-1},
\ens
where we require $x^{vac}_{t-1} \leq R_t$.  We still have our decision of whether to observe the environment $x^{obs}_t$, so our decision variables are
\bns
x_t = (x^{inv}_t, x^{vac}_t, x^{obs}_t).
\ens

While we are at it, we might as well include information about the weather such as temperature $I^{temp}_t$ and humidity $I^{hum}_t$ which can contribute to the spread of the flu.  We would model these in our ``other information'' variable
\bns
I_t &=& (I^{temp}_t,I^{hum}_t).
\ens
Our state variable becomes
\bn
S_t = \big(R_t, (I^{temp}_t,I^{hum}_t), (\mubar_t, \beta_t)\big).\label{eq:stateresourceconstraint}
\en
We now have a combination of a controllable physical state $R_t$ that we can observe perfectly, exogenous environmental information $I^t = (I^{temp}_t, I^{hum}_t)$, and the belief state $B_t = (\mubar_t, \beta_t)$ which captures our distribution of belief about the controllable state $\mu_t$ that we cannot observe.

Note how quickly our solvable two-dimensional problem just became a much larger five-dimensional problem.  This is a big issue if we are trying to use Bellman's equation, but only if we are using a lookup table representation of the value function (otherwise, we do not care).

\subsubsection{A spatial model}
Imagine that we have to allocate our supply of flu vaccines over a set of regions $\Ical$.  For this problem, we have a truth $\mu_{ti}$ and belief $(\mubar_{ti}, \beta_{ti})$ for each region $i\in\Ical$.  Next assume that $x^{vac}_{ti}$ is the number of vaccines allocated to region $i$, which is subject to the constraint
\bn
\sum_{i\in\Ical} x^{vac}_{ti} \leq R_{t}.\label{eq:spatialinventoryflu}
\en
Our inventory $R_t$ now evolves according to
\bns
R_{t+1} = R_t + x^{inv}_t - \sum_{i\in\Ical} x^{vac}_{ti}.
\ens
The coupling constraint \eqref{eq:spatialinventoryflu} prevents us from solving for each region independently.  This produces the state variable
\bn
S_t = \big(R_t, (\mubar_{ti}, \beta_{ti})_{i\in\Ical}\big). \label{eq:spatialstateresourceconstraint}
\en
What we have done with this extension is to create a state variable that is potentially very high dimensional, since spatial problems may easily range from hundreds to thousands of regions.


\subsubsection{Notes}
The problems in this section were chosen to bring out a series of modeling issues.  We made the transition from learning a static parameter $\mu$ to learning a dynamic parameter $\mu_t$ with no drift, and then a problem with an unknown drift.  All three of these problems involved state variables that described our beliefs about unknown parameters.

Then, we made the truth $\mu_t$ controllable, which is when we found that we could model the system where the state was $\mu_t$ (which is not observable), or the belief about $\mu_t$.
This problem falls in the domain of the POMDP community, where $\mu_t$ would be the state of our system, and then we have a belief about this state.

We then closed with two problems that combined the controllable but unobservable parameter $\mu_t$, along with controllable but observable parameter $R_t$.

We are now going to describe how the POMDP community approaches problems with controllable, but unobservable, states.  After this, we are going to present a  perspective that introduces a fresh way of thinking about these problems that opens up new solution approaches that draw on the four classes of policies described in section \ref{sec:designingpolicies}.

\subsection{The POMDP perspective}
The POMDP community approaches the controllable version of our flu problem by viewing it as a dynamic program with state $\mu_t$ and action $x_t$ that controls (or at least influences) this state.  Viewed from this perspective, $\mu_t$ is {\it the} state of the system.  Any reference to ``the state'' refers to the current value of $\mu_t$.  In our resource constrained system, we would add $R_t$ to the state variable giving us $S_t = (R_t, \mu_t)$, but for now we are going to focus on the unconstrained problem.

The community then shifts to the idea of modeling the belief about $\mu_t$, and then introduces the ``belief MDP'' where the belief is the state (instead of $\mu_t$).

The problem with these two versions of a Markov decision process is that there is not a clear model of who knows what. There is also the confusion of a ``state'' $s$ (sometimes called the physical state), and our belief $b(s)$ giving the probability that we are in state $s$, where $b(s)$ is its own state variable!! This issue arises not just in who has access to the value of $\mu_t$, but also information about the transition function.  In section \ref{sec:twoagentpomdp}, we are going to offer a new model that resolves the confusion about these two perspectives.

To help us present the POMDP perspective, we are going to make three assumptions:
\begin{description}
  \item[A1] Our state space (that is, the possible values of $\mu_t$) is discrete, which means we can write $S_t\in\Scal = \{s_1, \ldots, s_K\}$.  Note that this state space can become quite large when we have more than one dimension, as occurred with our other models (think about our spatially distributed problem).
  \item[A2] We are solving the problem in steady state.
  \item[A3] We can compute the one-step transition matrix $p(s'|s,x)$ which is the probability that we transition to $S=s'$ given that we are in state $s$ and take action $x$.  It is important to remember that $p(s'|s,x)$ is computed using
      \bns
      p(s'|s,x) & = & \E_S \E_{W|S}\{\mathbbm{1}_{\{S_{t+1}=s'=S^M(s, x, W)\}}|S_t=s\}.
      \ens
      The first expectation $\E_S$ captures our uncertainty about the state $S$, while the second expectation $\E_{W|S}$ captures the noise in the observation of $S$.  Computing the one-step transition matrix $p(s'|s,x)$ means we need to know both the transition function $S^M(s,x,W)$ and the probability distribution for $W$.
\end{description}
Table \ref{tab:POMDPnotation} presents the notation we use in our model, which parallels the notation for our original model.  The structure of our model, however, is completely standard.  Our model follows the presentation in \cite{Ross2008a}.
\begin{table}[tb]
\begin{center}
{\small
\begin{tabular}{|c|l|} \hline
\multicolumn{2}{|c|}{Physical (unobservable) system}\\ \hline
$S_t=s=\mu_t$    & Physical (unobservable) state $s$ (e.g. $\mu_t$) \\
$x_t$            & Decision (made by the controller) that acts on $S_t$ \\
$W_{t+1}$        & Exogenous information impacting the physical state $S_t$ \\
$S^M(s,x,w)$     & Transition function for physical state $S_t=s$ given $x_t=x$ and $W_{t+1}=w$ \\
$p(s'|s,x)$      & $Prob[S_{t+1} = s'|S_t=s,x_t=x]$ \\ \hline
\multicolumn{2}{|c|}{Controller system} \\ \hline
$b(s)$           & Probability (belief) we are in physical (unobservable) state $s$ \\
$W^{obs}$        & Noisy observation of physical state $S_t=s$ \\
$\Wcal^{obs}$    & Space of outcomes of $W^{obs}$ \\
$P^{obs}(w|s)$   & Probability of observing $W^{obs}=w$ given $S_t=s$ \\ \hline
\end{tabular}
}
\end{center}
\caption{Table of notation for POMDPs}
\label{tab:POMDPnotation}
\end{table}

With these assumptions, we can formulate the familiar form of Bellman's equations for discrete states and actions
\bn
V(s) = \max_x\left(C(s,x) + \sum_{s'\in\Scal} p(s'|s,x) V(s')\right). \label{eq:bellmanflusteadystate}
\en

The problem with solving Bellman's equation \eqref{eq:bellmanflusteadystate} to determine actions is that the controller determining $x$ is not able to see the state $s$.  The POMDP community addresses this by creating a belief $b(s)$ for each state $s\in\Scal$.  At any point in time, we can only be in one state, which means
\bns
\sum_{s\in\Scal} b(s) = 1.
\ens
The POMDP literature then creates what is known as the {\it belief MDP} in terms of the belief state vector $b = (b(s_1), \ldots, b(s_K)) = (b_1, \ldots, b_K)$. This is a dynamic program whose state is given by the continuous vector $b$.  We next introduce the transition function for the belief vector $b$ given by
\bns
B^M(b,x,W) &=& \textwrap{The transition function that gives the probability vector $b' = B^M(b,x,W)$ when the current belief vector (the prior) is $b$, we make decision $x$ and then observe the random variable $W$, which is a noisy observation of $\mu_t$ if we have chosen to make an observation.}
\ens
The function $B^M(b,x,W)$ returns a vector $b'$ that has an element $b'(s)$ for each physical state $s$.  We will write $B^M(b,x,W)(s)$ to refer to element $s$ of the vector returned by $B^M(b,x,W)$.

The belief transition function is an exercise in Bayes' theorem.  Let $b_t(s)$ be the probability we are in state $S_t = s'$ (this is our prior) at time $t$.  We assume we have access to the distribution
\bns
P^{obs}(w^{obs}|s) &=& \textwrap{the probability that we observe $W^{obs}=w^{obs}$ if we are in state $s$.}
\ens
Assume we are in state $S_t=s$ and take action $x_t$ and observe $W^{obs}_{t+1}=w^{obs}$.  The updated belief distribution $b_{t+1}(s')$ would then be
{\footnotesize
\bn
b_{t+1}(s'|b_t, x_t, W^{obs}_{t+1}=w^{obs}) \hspace{-.1in}&=&\hspace{-.1in} B^M(b_t,x,W^{obs}_{t+1}=w^{obs})(s')\nonumber\\
                   \hspace{-.1in}&=&\hspace{-.1in} Prob[S_{t+1}=s'|b_t, x_t, W^{obs}_{t+1}=w^{obs}] \nonumber \\
                   \hspace{-.1in}&=&\hspace{-.1in} \frac{Prob[W^{obs}_{t+1}=w^{obs}|b_t, x_t=x,S_{t+1}=s'] Prob[S_{t+1}=s'|b_t,x_t]}{Prob[W^{obs}=w^{obs}|b_t,x_t]} \label{eq:bayesupdatepomdp1}\\
                   \hspace{-.1in}&=&\hspace{-.1in} \frac{Prob[W^{obs}_{t+1}=w^{obs}|b_t, x_t,S_{t+1}=s'] \sum_{s\in\Scal} Prob[S_{t+1}=s'|S_t = s,b_t,x_t]Prob[S_t=s|b_t,x_t] }{Prob[W^{obs}_{t+1}=w^{obs}|b_t,x_t]} \nonumber \\
                   & & \label{eq:bayesupdatepomdp2}\\
                   \hspace{-.1in}&=&\hspace{-.1in} \frac{Prob[W^{obs}_{t+1}=w^{obs}|S_{t+1}=s'] \sum_{s\in\Scal}  Prob[S_{t+1}=s'|S_t = s,x_t] b_t(s)}{Prob[W^{obs}_{t+1}=w^{obs}|b_t,x_t]} \label{eq:bayesupdatepomdp3}\\
                   \hspace{-.1in}&=&\hspace{-.1in} \frac{P^{obs}(w^{obs}|s') \sum_{s\in\Scal}  P(s'|s,x_t)b_t(s)}{P^{obs}(w^{obs}|b_t,x_t)} \label{eq:bayesupdatepomdp4}\\
\en
}
Equation \eqref{eq:bayesupdatepomdp1} is a straightforward application of Bayes' theorem, where all probabilities are conditioned on the decision $x_t$ and the prior $b_t$ (which has the effect of capturing history).  Equation \eqref{eq:bayesupdatepomdp2} handles the transition from conditioning on the belief $b(s)$ that $S_t = s$, to the state $S_{t+1}=s'$ from which observations are made.  The remaining equations reduce \eqref{eq:bayesupdatepomdp2} by recognizing when conditioning on $b_t(s)$ does not matter, and substituting in the names of the variables for the differenet probabilities.

We compute the denominator in equation \eqref{eq:bayesupdatepomdp1} using
{\footnotesize
\bn
Prob[W^{obs}_{t+1}=w^{obs}|b_t,x_t] &=& \sum_{s'\in\Scal} Prob[W^{obs}_{t+1}=w^{obs}|S_{t+1}=s'] \sum_{s\in\Scal}Prob[S_{t+1}=s'|S_t=s,x_t] Prob[S_t=s|b_t,x_t] \\
                                    &=& \sum_{s'\in\Scal} P^{obs}(w^{obs}|s') \sum_{s\in\Scal} P(s'|s,x_t) b_t(s). \label{eq:bayesupdatepomdp4a}
\en
}
Equation \eqref{eq:bayesupdatepomdp4} is fairly straightforward to compute as long as the state space $\Scal$ is not too large (actually, it has to be fairly small), the observation probability distribution $P^{obs}(w^{obs}|S=s,x)$ is known, and the one-step transition matrix $P(s'|s,x)$ is known. Knowledge of $P^{obs}(w^{obs}|S=s,x)$ requires an understanding of the structure of the process of observing the unknown system.  For example, if we are sampling the population to learn about who has the flu, we might use a binomial sampling distribution to capture the probability that we sample someone with the flu.  Knowledge of the one-step transition matrix $P(s'|s,x)$, of course, requires an understanding of the underlying dynamics of the physical system.

This said, note that we have three summations over the state space to compute a single value of $b_{t+1}(s')$.  This has to be repeated for each $s'\in\Scal$, and it has to be computed for each action $x_t$ and observation $W^{obs}_{t+1}$.  That is a lot of nested loops.  The problem is that we are modeling two transitions: the evolution of the state $S_t$, and the evolution of the belief vector $b_t(s)$.  This would not be an issue if we were just simulating the two systems.  Equations \eqref{eq:bayesupdatepomdp4} and \eqref{eq:bayesupdatepomdp4a} require computing expectations to find the transition probabilities for both the physical state $S_t$ and the belief state $b_t$.

The POMDP community then approaches solving this dynamic program through Bellman's equation (for steady state problems) that can be written
\bns
V(b_t) &=& \max_x\big(C(b_t,x) + \E\{V(B^M(b_t,x,W^{obs}_{t+1}))|b_t,x\}\big).
\ens
It is better to expand the expectation operator over the actual random variables that are involved.  Assume that we are in physical state $S_t$ with belief vector $b_t$.  The expectation would then be written
\bn
V(b_t) &=& \max_x\big(C(b_t,x) + \E_{S_t|b_t} \E_{S_{t+1}|S_t} \E_{W^{obs}_{t+1}|S_{t+1}}\{V(B^M(b_t,x,W^{obs}))|b_t,x\}\big). \label{eq:bellmanflubeliefmdp5}
\en
Here, $\E_{S_t|b_t}$ integrates over the state space for $S_t$ using the belief distribution $b_t(s)$. $\E_{S_{t+1}|S_t}$ takes the expectation over $S_{t+1}$ given $S_t$.  Finally,  $\E_{W^{obs}_{t+1}|S_{t+1}}$ integrates over the space of observations given we are in state $S_{t+1}$.  These expectations would be computed using
{\small
\bn
V(b_t) &=& \max_x\left(C(b_t,x) + \sum_{s\in\Scal} b_t(s) \sum_{s'\in\Scal} p(s'|s,x) \sum_{w^{obs} \in\Wcal^{obs}} P^{obs}(w^{obs}|s,x) V(B^M(b_t,x,w^{obs}))\right). \label{eq:bellmanflubeliefmdp6}
\en
}
If equation \eqref{eq:bellmanflubeliefmdp6} can be solved, then the policy for making decisions for the controller is given by
\bns
X^*(b_t) &=& \argmax_x\left(C(b_t,x) + \sum_{s\in\Scal} b_t(s) \sum_{w^{obs} \in\Wcal^{obs}} P^{obs}(w^{obs}|s,x) V(B^M(b_t,x,W^{obs}))\right).
\ens

As if the computations behind these equations were not daunting enough, we need to also realize that we are combining the decisions of the controller with a knowledge of the dynamics (captured by the transition matrix) of the physical system.  The one-step transition matrix requires knowledge of both the transition function $S^M(s,x,W)$, which will not always be known to the controller.  We are going to illustrate a setting where the transition is not known to the controller below.


A challenge here is that even through we have discretized the unobservable state ($\mu$ for this problem), the vector $b$ is continuous.  However, Bellman's equation using the state $b$ has some nice properties that the research community has exploited.  Just the same, it is still limited to problems where the state space $\Scal$ of the unobservable system is relatively small.  Keep in mind that small problems can easily produce state spaces of 10,000, and we could never execute these equations with a state space that large.

We need to emphasize that while the POMDP community typically turns to Bellman's equation to solve a sequential decision problem, in practice this will rarely be computationally feasible.  This realization has motivated the development of a series of approximation strategies which are summarized in \cite{Ross2008a}.  One class of approximations that has attracted considerable attention is to restrict belief states to a sampled set $\Bhat$.

We urge readers to formulate the problem in terms of the canonical model given in section \ref{sec:modelingsequentialdecisions}, and then to consider all four classes of policies described in section \ref{sec:designingpolicies}.  Note that all four classes of policies are widely used for sequential decision problems, including those without a belief state (such as resource allocation problems), pure learning problems (where the only state variable is the belief state), as well as hybrids, such as the resource constrained problems.  Section \ref{sec:designingpolicyflu} sketches examples of each of the four classes of policies, none of which suffer from the curses of dimensionality that we encounter in equation \eqref{eq:bayesupdatepomdp4}.

\section{A two-agent model of the flu application}
\label{sec:twoagentpomdp}
We now offer a completely different approach for modeling POMDPs using the context of our flu application.  We start by presenting a two-agent model of the POMDP for the flu problem.  We then turn our attention to knowledge of the transition function itself, rather than just parameters.

\subsection{A two-agent formulation of the POMDP}
There are two perspectives that we can take in any POMDP: one from the perspective of the environment, and one from the perspective of the controller that makes decisions:
\begin{description}
  \item[The environment perspective] The environment (sometimes called the ``ground truth'') knows $\mu_t$, but cannot make any decisions (nor does it do any learning).
  \item[The controller perspective] The controller makes decisions that affect the environment, but is not able to see $\mu_t$.  Instead, the controller only has access to the belief about $\mu_t$.
\end{description}
\begin{figure}[t]
\horizontalline
{\dense{\small
\begin{description}
  \item[State variables] $S^{env}_t = \big(\mu_t,\delta\big)$ (we include the drift $\delta$, even if it is not changing).
  \item[Decision variables] There are no decisions.
  \item[Exogenous information] $W^{env}_{t+1} = \varepsilon^\mu_{t+1}$.
  \item[Transition function] $S^{env} = S^{M,env}(S^{env}_t,,W^{env}_{t+1})$, which includes equation \eqref{eq:timevaryingmodel} describing the evolution of $\mu_t$.
  \item[Objective function] Since there are no decisions, we do not have an objective function.
\end{description}
\horizontalline
\caption{The canonical model of the environment.}
\label{fig:environmentagent}
}}
\end{figure}
The model of the environment agent is given in figure \ref{fig:environmentagent}.  The model of the controlling agent is given in figure \ref{fig:controllingagent}.

\begin{figure}[t]
\horizontalline
{\dense{\small
\begin{description}
  \item[State variables] $S^{cont}_t = \big((\mubar_t, \beta_t), (\deltabar_t,\beta^\delta_t)\big)$.
  \item[Decision variables] $x_t = (x^{vac}_t,x^{obs}_t)$.
  \item[Exogenous information] $W^{cont}_{t+1}$ which is our noisy estimate of how many people have the flu (and we only obtain this if $x^{obs}_t = 1$).
  \item[Transition function] $S^{cont}_{t+1} = S^{M,cont}(S^{cont}_t,x_t,W^{cont}_{t+1})$, which consist of equations \eqref{eq:learningtransitionmubarflu} and \eqref{eq:learningtransitionsigmabarflu}.
  \item[Objective function] We can write this in different ways. Assuming we are implementing this in a field situation, we want to optimize cumulative reward.  Let:
      \bns
      c^{obs} &=& \textwrap{The cost of sampling the population to estimate the number of people infected with the fly,}\\
      C^{vac}(\mubar_t) &=& \textwrap{the cost we assess when we think that the number of infected people is $\mubar_t$.}
      \ens
      Now let $C^{cont}(S_t,x_t) = c^{obs}x^{obs}_t + C^{vac}(\mubar_t)$ be the cost at time $t$ when we are in state $S_t$ and make decision $x_t$ (note that $x^{vac}_t$ impacts $S_{t+1}$).  Finally, we want to optimize
      \bn
      \max_\pi \E\left\{\sum_{t=0}^T C^{cont}(S_t,X^\pi_t(S_t)) | S_0\right\}. \label{eq:objectiveflu}
      \en
\end{description}
\horizontalline
\caption{The canonical model of the controlling agent.}
\label{fig:controllingagent}
}}
\end{figure}

It is best to think of the two perspectives as agents, each working in their own world.  There is the ``environment agent'' which does not make decisions, and the ``controlling agent'' which makes decisions, and performs learning about the environment that cannot be observed (such as $\mu_t$).  Once we have identified our two agents, we need to define what is known by each agent.  This begins with who knows what about parameters such as $\mu_t$, but it does not stop there.

Table \ref{tab:statevariables} shows the environmental state and controlling state variables for each of the variations of our flu problem that we presented in section \ref{sec:fluvariations}.  A few observations are useful:
\begin{table}[tb]
\begin{center}
{\small
\begin{tabular}{|c|c|c|l|} \hline
 & $S^{env}_t$ & $S^{cont}_t$ & Description \\ \hline
1) &
$(\mu_t)$ &
$(\mubar_t, \beta_t)$ &
Static, unknown truth \\
2) &
$\big((\mu_t), (I^{temp}_t,I^{hum}_t)\big)$ &
$\big(R_t, (I^{temp}_t,I^{hum}_t), (\mubar_t, \beta_t)\big)$ &
Resource constrained with exogenous information   \\
3) &
$\big(\mu_t,\delta\big)$ &
$\big((\mubar_t, \beta_t), (\deltabar_t,\beta^\delta_t)\big)$ &
Dynamic model with uncertain drift  \\
%
%
4) &
$\big(\mu_t, x^{vac}_{t-1}, \theta^{vac}\big)$ &
$\big(\mubar_t, \beta_t\big)$ &
Dynamic model with a controllable truth \\
5) &
$\big((\mu_{ti})_{i\in\Ical}, x^{vac}_{t-1}, \theta^{vac}\big)$ &
$\big(R_t, (\mubar_{t}, \beta_{t})\big)$ &
Resource constrained model \\
6) &
$\big((\mu_{ti})_{i\in\Ical}, x^{vac}_{t-1}, \theta^{vac}\big)$ &
$\big(R_t, (\mubar_{ti}, \beta_{ti})_{i\in\Ical}\big)$ &
Spatially distributed model \\ \hline
\end{tabular}
}
\end{center}
\caption{Environmental state variables and controller state variables for different models.}
\label{tab:statevariables}
\end{table}

\begin{itemize}
  \item The two-agent perspective means we have two systems.  The environment agent is a simple system with no decisions, but with access to $\mu_t$ and the dynamics of how vaccinations affect $\mu_t$.  The state of the system for the environment agent is $S^{env}_t$ which includes $\mu_t$.  The state for the system for the controlling agent, $S^{cont}_t$, is the belief about $\mu_t$, along with any other information known to the controlling agent such as $R_t$.  The two systems are completely distinct, beyond the ability to communicate.
  \item In model 2, we model the temperature $I^{temp}_t$ and humidity $I^{hum}_t$ as state variables for both the environment, which presumably would control changes to these variables, and the controlling agent, since we have assumed that the controlling agent is able to observe these perfectly.  We could, of course, insist that the controller can only observe these through imperfect instruments, in which case they would be handled in the same way we handle $\mu_t$.
  \item Normally a state variable $S_t$ should only include information that changes over time (otherwise the information would go in the initial state $S_0$).  For this presentation, we included information such as the drift $\delta$ (model 3) and the effect of vaccinations on the prevalence of the flu $\theta^{vac}$ (model 4) in the environmental state variables to indicate information known to the environment but not to the controlling agent.
  \item In model 4, we include the decision $x^{vac}_{t-1}$ in the state variable for the environment.  We assume that the controlling agent makes the decision to vaccinate $x^{vac}_{t-1}$ at time $t-1$ which is then communicated to the environment (which is how it gets added to $S^{env}_t$) and is then implemented during time period $t$.  The information arrives to the environment through the exogenous information variable $W^{env}_t$.
  \item For models 5 and 6, we see how quickly we can go from two or three dimensions, to hundreds or thousands of dimensions.  The spatially distributed model cannot be solved using standard discrete representations of state spaces, but approximate dynamic programming has been used for very high-dimensional resource allocation problems (see \cite{SiDaGe09} and \cite{Bouzaiene-Ayari2016}).
\end{itemize}

In addition to modeling what each agent knows, we have to model communication.  This will become an important issue when we model multiple controlling agents which we address in section \ref{sec:multiagent}.  For our problem with a single controlling agent and a passive environment, there are only two types of communication: 1) the ability of the controlling agent to observe the environment (with noise) and 2) the communication of the decision $x^{vac}_t$ to the environment.

It is not hard to see that {\it any} learning problem can (and we claim should) be presented using this ``two-agent'' perspective.

Below we are going to illustrate a setting where the controlling agent does not know the transition function.  When we move to multiagent systems, we will add the dimension of learning the policies of other agents.

\subsection{Transition functions for two-agent model}

Our two-agent model has focused on what each agent knows (the state variable), but there is another dimension that deserves a closer look, which is the transition function.  Assume that the true model describing the evolution of $\mu_t$ (known only to the environment) is

{\small
\bn
\mu_{t+1} = \theta^\mu_{0}\mu_t + \theta^\mu_{24} \mu_{t-24} + (\theta^{temp}_0 U_t+\theta^{temp}_1 U_{t-1} +\theta^{temp}_2 U_{t-2}) - (\theta^{vac}_1 x^{vac}_{t-1} + \theta^{vac}_2 (x^{vac}_{t-1})^2) + \varepsilon^\mu_{t+1}. \label{eq:flucontrollabletruthenv}
\en
}
where
\bns
U_t = \big(\max\{0,I^{temp}_t - I^{threshold}\}\big)^2
\ens
where $I^{threshold}$ is a threshold temperature (say, 25 degrees F) below which colds and sneezing begins to spread the flu.  The inclusion of temperature over the current and two previous time periods captures lag in the onset of the flu due to cold temperatures.

For certain classes of policies, the controlling agent needs to develop its own model of the evolution of the flu.  The controlling agent would not know the true dynamics in equation \eqref{eq:flucontrollabletruthenv} and might instead use the following time-series model for the observed number of flu cases $W_t$:
\bn
W_{t+1} = \theta^W_0 W_t + \theta^W_1 W_{t-1} + \theta^W_2 W_{t-2} - \theta^{vac} x^{vac}_{t-1}+ \varepsilon^W_{t+1}. \label{eq:flucontrollabletruthcont}
\en
Our model in equation \eqref{eq:flucontrollabletruthcont} is a reasonable time-series model for the sequence of observations $W_1, \ldots, W_t$ to predict $W_{t+1}$.  There are, however, several errors in this model:
\begin{itemize}
  \item The controlling agent is using observations $W_t, W_{t-1}$ and $W_{t-2}$ while the environment uses $\mu_t$, which is not observable to the POMDP.
  \item The controlling agent did not realize there was a 24-hour lag in the development of the flu.
  \item The controlling agent is ignoring the effect of temperature.
  \item The controlling agent is not properly capturing the effect of vaccinations on infections.
\end{itemize}
Just the same, ignorance is bliss and our controlling agent moves forward with his best effort at modeling the evolution of the flu.  Assume that the model is a reasonable fit of the data.  We suspect that a careful examination of the errors (they should be independent and identically distributed) might fail a proper statistical test, but it is also possible that we cannot reject the hypothesis that the errors do satisfy the appropriate conditions. This does not mean that the model is correct - it just means that we do not have the data to reject it.

Now imagine that a graduate student is writing a simulator for the flu model, and assume that there is only one person writing the code (which is typically what happens in practice).  Our erstwhile graduate student will create the true transition equation \eqref{eq:flucontrollabletruthenv}.  When she goes to create the transition model used by the controller, she would create the best approximation possible given the information she was allowed to use, but she would know immediately that there are a number of errors in her approximation.  This would allow her to declare that this model is ``non-Markovian,'' but it is only because she is using her knowledge of the true model.


It is useful to repeat the famous quote of George Box that ``all models are wrong, but some are useful.''  We suspect that most (if not all) of the errors that are inherent in any statistical model would allow us to show, given enough data, that the errors $\varepsilon_t$ are not independent across time, although the errors may be small enough that, given our dataset, we cannot reject the hypothesis that they are independent.  This supports our second claim
\begin{changemargin}{1.0cm}{1.0cm}
{\it
Claim 2: All models of real problems are (possibly) non-Markovian.
}
\end{changemargin}
While Claim 2 seems to contradict Claim 1 (all properly modeled systems are Markovian), the conflict boils down to the interpretation of the model.  When we claim that all properly modeled systems are Markovian (such as equation \eqref{eq:flucontrollabletruthcont}), we were addressing the tendency of some in the community to represent $W_t$ as the ``state'' of our process, when in fact the real state for our assumed model for the controller is $(W_t, W_{t-1}, W_{t-2}, x^{vac}_t)$ (some would say that this is history dependent).

The observation that the controller model \eqref{eq:flucontrollabletruthcont} is non-Markovian, on the other hand, arises only because the modeler is able to cheat and see the true model, which would never happen in a real system.  This is why the two-agent formulation is important: We need to create a model of the environment that is known only to the environment.  This includes not only the value of variables such as $\mu_t$, but also the structure of the system model \eqref{eq:flucontrollabletruthenv}.

Thus, given the true dynamics in \eqref{eq:flucontrollabletruthenv}, it is possible to insist that the controlling agent's model of the environment in \eqref{eq:flucontrollabletruthcont} is non-Markovian, but this statement uses information that the controlling agent would never have in practice.  {\it This means that the model in \eqref{eq:flucontrollabletruthcont} is Markovian not because it is truly Markovian, but because we assume it is.}

Recognizing the difference between the truth and a model raises the philosophical question of what we mean by statements such as ``the model is Markovian'' or ``the solution is optimal.''  We suspect that few would disagree with the position that these statements only make sense relative to a {\it model} of a real problem, rather than the real problem itself.  Thus, if we are given an observation of, say, prices $p_0, p_1, \ldots, p_t$, it would not make any sense to say whether this series is Markovian.  We would have to create a model of the process, just as we did for the evolution of the flu in equation \eqref{eq:flucontrollabletruthcont}, and then work with this model.  However, it is possible to train policies using historical data, but this is just what we are doing with our time series equation \eqref{eq:flucontrollabletruthcont}.  The trained model is just an approximation, which means that even if we could find an optimal policy, it is only optimal relative to the approximation.  We ask the reader to keep this in mind in the next section when we transition to designing policies.

\section{Designing policies for the flu problem}
\label{sec:designingpolicyflu}
Once we formulate our models of each agent, we need to design policies for the controlling agent.  The creation of effective, high quality policies can be major projects. What we want to do is to sketch examples of each of the four classes of policies to help reinforce why it is important to understand all four classes.

\subsection{Policy function approximations}
Policy function approximations are analytic functions that map states to actions.  Of the four classes of policies, this is the only class that does not involve an imbedded optimization problem.

For our flu problem, it is common to use the structure of the problem to identify simple functions for making decisions.  For example, we might use the following rule for determining whether to make an observation of the environment:
\bn
X^{pfa-obs}(S_t|\theta^{obs}) = \left\{\begin{tabular}{cl} 1 & $\sigmabar_t/\mubar_t \geq \theta^{obs}$ \\
                                 0 & \mbox{Otherwise.}  \end{tabular}\right. \label{eq:pfa-obs}
\en
The policy captures the intuition that we want to make an observation when the level of uncertainty (captured by the standard deviation of our estimate of the true prevalence), relative to the mean, is over some number.  The parameter $\theta^{obs}$ has to be tuned, which we do using the objective function \eqref{eq:objectivestaticflu}.  A nice feature of the tunable parameter is that it is unitless.

To determine $x^{vac}_t$, we might set $\mu^{vac}$ as a target infection level, and then vaccinate at a level that we believe (or hope?) that we get down to the target.  To do this, first compute
\bns
\zeta_t(\theta^{vac}) = \frac{1}{\theta^{vac}}\max\{0,\big(\mubar_t - \mu^{vac}\big)\}.
\ens
We can view $\zeta_t$ as the distance to our goal $\mu^{vac}$.  This calculation ignores the uncertainty in our estimate $\mubar_t$, so instead we might want to use
\bns
\zeta_t(\theta^\zeta) = \max\{0,\big(\mubar_t + \theta^\zeta \sigmabar_t - \mu^{vac}\big)\}.
\ens
This policy is saying that $\mu_t$ {\it might} be as large as $\mubar_t + \theta^\zeta \sigmabar_t$, where $\theta^\zeta$ is a tunable parameter. Now our policy for $x^{vac}$ would be be
\bn
X^{pfa-vac}(S_t|\theta^{vac},\theta^\zeta) = \frac{1}{\theta^{vac}}\zeta_t(\theta^\zeta). \label{eq:pfa-vac}
\en

Using our policy $X^{obs}(S_t)$, we can write our policy for $x_t = (x^{vac}_t,x^{objs}_t)$ as
\bns
X^{PFA}(S_t|\theta) = \big(X^{pfa-vac}(S_t|\theta^{vac},\theta^\zeta), X^{pfa-obs}(S_t|\theta^{obs})\big).
\ens
where $\theta = (\theta^{vac},\theta^{obs},\theta^\zeta)$.  This policy would have to be tuned in the objective function \eqref{eq:objectiveflu}.  This policy could then be compared to that obtained by approximating Bellman's equation.

An alternative approach for designing a policy function approximation is to assume that it is represented by a linear model
\bns
X^{PFA}(S_t|\theta) = \sum_{f\in\Fcal} \theta_f \phi_f(S_t).
\ens
Parametric functions are easy to estimate, but they require that we have some intuition into the structure of the policy.  An alternative is to use a neural network, where $\theta$ is the weights on the links in the graph of the neural network.  It is important to keep in mind that neural networks tend to be very high dimensional ($\theta$ may have thousands of dimensions), and they may not replicate obvious properties.  Either way, we would tune $\theta$ using the objective function in \eqref{eq:objectiveflu}.

\subsection{Cost function approximations}
We are going to illustrate CFAs using the spatially distributed flu vaccination problem, where we assume we are allowed to observe just one region $x\in\Ical$ at a time (we have just one inspection team).  Assume that we can only treat one region at a time, where we are always going to treat the region that has the highest estimated prevalence of the flu.

We do not know $\mu_{tx}$, but at time $t$ assume that we have an estimate $\mubar_{tx}$ for the prevalence of the flu in region $x\in\Ical$, where we assume that $\mu_x \sim N(\mubar_{tx}, \sigmabar^{2}_{tx})$.  We use this belief to decide which region to vaccinate, which we describe using the policy
\bns
X^{vac}(S_t) = \argmax_{x\in\Ical} \mubar_{tx}.
\ens

We then have to decide which region to observe.  We can approach this problem as a multiarmed bandit problem, where we have to decide which region (``arm'' in bandit lingo) to observe.  The most popular class of policies for learning problems in the computer science community is known as upper confidence bounding for multiarmed bandit problems.  A class of UCB policy is interval estimation which would choose the region $x$ that solves
\bn
X^{obs-IE}(S_t|\theta^{IE}) = \argmax_{x\in\Ical} \big(\mubar_{tx} + \theta^{IE} \sigmabar_{tx}\big)
\en
where $\sigmabar_{tx}$ is the standard deviation of the estimate $\mubar_{tx}$.

The policy $X^{obs-IE}(S_t|\theta^{IE})$ is a form of parametric cost function approximation: it requires solving an imbedded optimization problem, and there is no explicit effort to approximate the impact of a decision now on the future.  It is easy to compute, but $\theta^{IE}$ has to be tuned.  To do this, we need an objective function.  Note that we are going to tune the policy in a simulator, which means we have access to $\mu_{tx}$ for all $x\in\Ical$.

Let $x^{obs}_t = X^{obs-IE}(S_t|\theta^{IE})$ be the region we choose to observe given what we know in $S_t$.  This gives us the observation
\bns
W_{t+1,x^{obs}_t} = \mu_{t,x^{obs}_t} + \varepsilon_{t+1},
\ens
where $\varepsilon \sim N(0,\sigma^2_W)$.  We would then use this observation to update the estimates $\mubar_{t,x^{obs}_t}$ using the updating equations \eqref{eq:learningtransitionmubar} - \eqref{eq:learningtransitionsigmabar}. 

It is important to remember that the true prevalence $\mu_{tx}$ is changing over time as a result of our policy of observation and vaccination, so we are going to refer to it as $\mu^\pi_{tx}(\theta^{IE})$, where the observation policy is parameterized by $\theta^{IE}$.

We are learning in the field, which means we want to minimize the prevalence of the flu over all regions, over time.  Since we are using a simulator to evaluate policies, we would evaluate our performance using the true level of flu prevalence, given by
\bn
F^\pi(\theta^{IE}) = \E_{S_0} \left\{\sum_{t=0}^{T} \sum_{x\in\Ical} \mu^\pi_{tx}(\theta^{IE}) |S^0\right\}. \label{eq:spatialflupolicy}
\en
We then need to tune our policy by solving
\bns
\min_{\theta^{IE}} F^\pi(\theta^{IE}).
\ens

\subsection{Policies based on value functions}
Any sequential decision problem with a properly defined state variable can be solved using Bellman's equation:
\bns
V_t(S_t) = \max_x \big(C(S_t,x_t) + \E\{V_{t+1}(S_{t+1})|S_t,x_t\}\big),
\ens
which gives us the policy
\bns
X^{VFA}(S_t) = \argmax_x \big(C(S_t,x_t) + \E\{V_{t+1}(S_{t+1})|S_t,x_t\}\big).
\ens
In practice we cannot compute $V_t(S_t)$, so we resort to methods that approximate the value function, following one of the styles given in equations \eqref{eq:vfa1}-\eqref{eq:vfa4}.

The use of approximate value functions has been recognized for a wide range of dynamic programming and stochastic control problems.  However, it has been largely overlooked for problems with a belief state, with the notable exception of the literature on Gittins indices \citep{gittins2011}, which reduces high-dimensional belief states (the beliefs across an entire set of arms) down to a series of dynamic programs with one per arm.

In principle, approximate dynamic programming can be applied to even high-dimensional problems, including those with belief states, by replacing the value function $V_t(S_t)$ with a statistical model such as
\bns
V_t(S_t) \approx \Vbar_t(S_t|\theta) = \sum_{f\in\Fcal}\theta_f \phi_f(S_t),
\ens
where $(\phi_f(S_t))_{f\in\Fcal}$ is a set of features.  Alternatively, we might approximate $\Vbar_t(S_t)$ using a neural network.

We note that we might write our policy as
\bn
X^{VFA}(S_t|\theta) = \argmax_x \left(C(S_t,x) + \sum_{f\in\Fcal}\theta_f \phi_f(S_t,x)\right). \label{eq:tunedvfaflu}
\en
where $(\phi_f(S_t,x_t)_{f\in\Fcal}$ is a set of features involving both $S_t$ and $x_t$. For example, we might design something like
\bns
X^{VFA}(S_t|\theta) = \argmax_x \left(C(S_t,x) + \big(\theta_{t0} + \theta_{t1} \mubar_t + \theta_{t2} \mubar^2_t + \theta_{t3} \sigmabar_t + \theta_{t4} \beta_t \sigmabar_t\big)\right).
\ens
There are a variety of strategies for fitting $\theta$ that have been developed under the umbrella of approximate dynamic programming \citep{PowellADP2011} and reinforcement learning \citep{Sutton2018}.


\subsection{Direct lookahead policy}
Direct lookahead policies involve solving an approximate lookahead model that we previously gave in equation \eqref{eq:optimalpolicyLAapproxpi}, but repeat it here for convenience:

{\small
\bn
\hspace{-0.20in} X^{DLA}_t(S_t) \hspace{-0.10in}&=&\hspace{-0.10in} \argmax_{x_t} \left(C(S_t,x_t) + \Etilde \left\{\max_{{\tilde \pi}} \Etilde \left\{\sum_{t'=t+1}^T C(\Stilde_{tt'},\Xtilde^{\tilde \pi}(\Stilde_{tt'})) | \Stilde_{t,t+1}\right\}|S_t,x_t\right\}\right). \label{eq:optimalpolicyLAapproxpi2}
\en
}
The problem with direct lookahead policies is that it requires solving a stochastic optimization problem (to solve the original stochastic optimization problem).  To make it tractable, we can introduce various approximations.  Some that are relevant for our problem setting could be:
\begin{description}
  \item[1)] Use a deterministic approximation.  These are effective for pure resource allocation problems (google maps using a deterministic lookahead to find the best path to the destination over a stochastic graph), but seem unlikely to work well for learning problems.
  \item[2)] Use a parameterized policy for ${\tilde \pi}$.  We could use any of the policies suggested above as our lookahead policy.  We would then also have to use Monte Carlo sampling to approximate the expectations.
  \item[3)] We can solve a simplified Markov decision process.
  \item[4)] We could approximate the lookahead using Monte Carlo tree search.
\end{description}

We are going to illustrate the third approach.  We start with model 3 of the flu problem, which requires the state variable
\bns
S^{cont}_t = \big((\mubar_t, \beta_t), (\deltabar_t,\beta^\delta_t)\big).
\ens
We might be able to do a reasonable job of solving a dynamic program with a two-dimensional state variable (using discretization), but not a four-dimensional state.  One approximation strategy is to fix the belief about the drift $\delta$ by holding $(\deltabar_t,\beta^\delta_t)$ constant.  This means that we continue to model the true $\delta$ with uncertainty, but we ignore the fact that we can continue to learn and update the belief.  This means the state variable $\Stilde_{tt'}$ in the lookahead model is given by
\bns
\Stilde_{tt'} = ({\tilde \mubar}_{tt'}, {\tilde \beta}_{tt'}).
\ens

Assuming that we can discretize the two-dimensional state, we can solve the lookahead model using classical backward dynamic programming on this approximate model (we could do this in steady state, or over a finite horizon, which makes more sense).  Solving this model will give us exact value functions $\Vtilde_{tt'}(\Stilde_{tt'})$ for our approximate lookahead model, from which we can then find the decision to make now given by
\bn
X^\pi_t(S_t) = \argmax_x \left(C(S_t,x) + \E \{ \Vtilde_{t,t+1}(\Stilde_{t,t+1})|S_t\}\right). \label{eq:tunedvfaflu}
\en
Then, we implement $x_t = X^\pi_t(S_t)$, step forward to $t+1$, observe $W_{t+1}$, update to state $S_{t+1}$ and repeat the process.

\subsection{A hybrid policy}
We have two types of decisions: whether to observe $x^{obs}_t$, and how many to vaccinate $x^{vac}_t$.  We can combine them into a single, two-dimensional decision $x_t = (x^{obs}_t,x^{vac}_t)$ and then think of enumerating all possible actions.  However, we can also use hybrids.  For example, we could use the policy function in equation \eqref{eq:pfa-obs}, but then turn to any of the other four classes of policies for $x^{vac}_t$.  This not only reduces the dimensionality of the problem, but might help if we feel that we have confidence in the function for $x^{obs}_t$ but are less confident designing a function for $x^{obs}_t$.

\subsection{Notes}
The POMDP literature focuses on who knows what about the parameter $\mu_t$, but it appears to completely overlook knowledge of the system model $S^M(S_t,x_t, W_{t+1})$.  We illustrated this by giving the true system model (known to the environment) in equation \eqref{eq:flucontrollabletruthenv}, and then described a plausible approximation created by the controller in equation \eqref{eq:flucontrollabletruthcont}.  The POMDP literature ignores this issue entirely, and implicitly assumes that the controller has access to the system model through its knowledge of the one-step transition matrix.

Both classes of lookahead policies (policies based on VFAs and DLAs) need an explicit model of the future.  The controller would have to use his own estimate of the transition function, rather than the true transition function known only to the environment.

The policies in the policy search class (the PFAs and CFAs) do not explicitly depend on the transition function, but both involve tunable parameters that have to be tuned.  If these are being tuned offline in a simulator, then this simulator would also have to use the approximate transition function known to the controller.

It appears that the only way to avoid using the controller's version of the transition function is to do online tuning of a PFA or CFA, which means we are doing tuning in the field.

\section{Multiagent systems}
\label{sec:multiagent}
We can extend our ``two-agent'' formulation of learning problems (with a controlling agent and an environment agent) to general multiagent problems with multiple controlling agents.  We begin by defining our different classes of agents:
\begin{itemize}
  \item The environment agent - This agent cannot make any decisions, or perform any learning (that is, anything that implies intelligence).  This is the agent that would know the truth about unknown parameters that we are trying to learn, or which performs the modeling of physical systems that are being observed by other agents.  Controlling agents are, however, able to change the environment.
  \item Controlling agents - These are agents that make decisions that act on other agents, or the environment agent.  Controlling agents may communicate information to other controlling and/or learning agents.  One form of control is when one agent wants to influence the decision that falls under the control of another agent.  This has to be done by creating an incentive for the other agent to choose a decision aligned with the wishes of the first agent.
  \item Learning agents - These agents do not make any decisions, but can observe and perform learning (about the ground truth and/or other controlling agents), and communicate beliefs to other agents.
  \item Combined controlling/learning agents - These agents perform learning through observations of the ground truth or other agents, as well as making decisions that act on the ground truth or other agents.
\end{itemize}
We are not requiring that controlling agents be homogeneous.  Each agent can control specific types of decisions, and can work at different levels (e.g. leader/follower).

We are going to model multiagent systems by simply extending the notation for our basic canonical model, just as we did for our two-agent POMDP.  In fact, we are going to use this model to represent each agent.  We do this using the following notation
\bns
\Qcal     &=& \textwrap{The set of agents, which we index by $q\in\Qcal$.}\\
S_{tq}    &=& \textwrap{The state variable for agent $q$ (this captures everything known by agent $q$) at time $t$.}\\
x_{tqq'}  &=& \textwrap{A decision made by agent $q$ that acts on agent $q'$ at time $t$.}\\
x_{tq}    &=& (x_{tqq'})_{q'\in\Qcal}\\
W_{t+1,qq'} &=& \textwrap{Information arriving to agent $q$ from agent $q'$ between $t$ and $t+1$. This can include the actions of other agents $q'$ acting on $q$.  These actions are communicated through $W_{t+1,qq'}$ and captured by $S_{t+1,q}$.}\\
W_{t+1,q} &=& (W_{t+1,qq'})_{q'\in\Qcal}.\\
C_q(S_{tq},x_{tq}) &=& \textwrap{Cost/contribution generated by agent $q$.}
\ens
Just as we created a new approach for modeling POMDPs using a two-agent formulation, we are going to propose that we can handle any multiagent system of arbitrary complexity by modeling the behavior of each individual agents using our standard modeling framework.  There are, of course, some issues that arise that we need to address.
\begin{itemize}
  \item Active observations - We saw this above with the decision $x^{obs}_t$ of whether to observe the environment, which we assume comes at a cost.  Now we add the dimension of actively observing other agents.  For example, a navy ship has to make the decision whether to turn on its radar to observe another ship, which simultaneously reveals the location of the ship sending the radar.
  \item Modeling communication - We have to model the act of sending information in $S_{tq}$ to another agent $q'$.  The information may be sent accurately, or with some combination of noise and bias.
  \item Receiving information - If information is sent from $q'$ to $q$, agent $q$ has to update her own beliefs, which has to reflect the confidence that agent $q$ has in the information coming from agent $q'$.
  \item Communication architecture - We have to decide who can communicate what to whom.  It will generally not be the case in more complex systems that any agent can (or would) send everything in their state vector $S_{tq}$ to every other agent.  We may have coordinating agents that communicate with everyone, make decisions and then send these decisions (in some form) to other agents.
\end{itemize}

An important dimension of multiple controlling agents is the formation of beliefs about the behavior of other agents.  This introduces the issue of {\it modeling} how other agents make decisions.  As with any statistical model, it will be an approximation, since we will have to formulate our belief about {\it how} another agent is making decisions.  This will likely be in the form of a parametric model with tunable parameters, which we will need to tune through a sequence of observations of the actions of other agents (who may be responding to our actions).  However, coming up with this model means that we have to assess {\it how} we think the other agent is making decisions (and how smart they may be).

We feel that these issues will make it difficult to have a discussion about optimal policies.  We can find an optimal policy for an agent given the assumed models of the transition functions and policies of other agents, but given the difference between assumed functions and true ones, an agent's optimal policy is not going to be optimal for the system.

A thorough treatment of multiagent systems is beyond the scope of this chapter, but we hope this hints at how our framework can be extended to more complex problems.

\clearpage
\singlespace
\addcontentsline{toc}{section}{References}
\bibliographystyle{agsm}

\end{document}